\documentclass[11pt]{article}
\usepackage[utf8]{inputenc}
\usepackage{graphicx,tabularx}
\usepackage[margin=2.5cm]{geometry} 
\usepackage{siunitx}
\usepackage{float}
\usepackage[table]{xcolor}
\usepackage[backend=biber,style=ieee,citestyle=ieee]{biblatex}
\usepackage{subcaption,amsfonts,graphicx,tabu,amsmath}
\usepackage{pdfpages}

\usepackage[scaled=1]{helvet} % helvetica
\usepackage[helvet]{sfmath} % also sans serif for maths

\usepackage{todonotes}
\usepackage{wrapfig}

\newcommand{\ts}[1]{_{\text{#1}}}

\usepackage{pgfplots}
\DeclareUnicodeCharacter{2212}{−}
\usepgfplotslibrary{groupplots,dateplot}
\usetikzlibrary{patterns,shapes.arrows}
\pgfplotsset{compat=newest}

\usepackage{booktabs,array}

\title{\LARGE \bf Predicting battery end of life from solar off-grid system field data using machine learning}

\author{Antti Aitio, David Howey}
\begin{document}

\maketitle

\noindent\rule[0.5ex]{\linewidth}{1pt}

\begin{abstract}
Hundreds of millions of people lack access to electricity. Decentralised solar-battery systems are key for addressing this  whilst avoiding carbon emissions and air pollution, but are hindered by relatively high costs and rural locations that inhibit timely preventative maintenance. Accurate diagnosis of battery health and prediction of end of life from operational data improves user experience and reduces costs. But lack of controlled validation tests and variable data quality mean existing lab-based techniques fail to work. We apply a scaleable probabilistic machine learning approach to diagnose health in 1027 solar-connected lead-acid batteries, each running for 400-760 days, totalling 620 million data rows. We demonstrate 73\% accurate prediction of end of life, eight weeks in advance, rising to 82\% at the point of failure. This work highlights the opportunity to estimate health from existing measurements using `big data' techniques, without additional equipment, extending lifetime and improving performance in real-world applications. 

\smallbreak
    
    
\end{abstract}

\noindent\rule[0.5ex]{\linewidth}{1pt}

\section{Introduction}

To achieve universal electricity access, the number of decentralised solar-battery systems and solar mini-grids in areas without grid access will need to increase ten-fold \cite{ESMAP2019EnergyMakers}, but this is inhibited by the relatively high costs and uncertain lifetimes of batteries \cite{Lee2018TheAfrica}. Diagnosis and prediction of battery state of health (SoH) in real-world operating environments is required for operational safety, warranties, and planning of maintenance, as well as for improving designs by understanding the impact of varying usage on battery life. Diagnosis in the field is challenging because direct measurement of SoH using standardised performance tests is usually not possible due to the costs of the required service interruption and testing equipment. Therefore health diagnosis should be performed directly from monitored operational data, for example battery terminal voltage, temperature and current. However, this means that the controlled operating conditions that would normally ensure consistent health estimation in laboratory tests are missing. Further complication arises because the most common battery health metrics---capacity and internal resistance---are influenced by operating conditions. Finally, the current, voltage and temperature sensors used in many real-world applications are not of the same accuracy as those in laboratories, and data recording is often incomplete, resulting in gaps in the time series. 

Techniques for battery health estimation can broadly be classified into model-based and data-driven methods \cite{Farmann2015CriticalVehicles,Berecibar2016,Xiong2018TowardsMethods,Li2019}. Model-based approaches typically use an electrical equivalent circuit model combined with techniques from feedback control to track internal states, such as state of charge (SoC), and parameters, such as resistance and capacity. Gradual evolution in the parameters enables state of health estimation using Bayesian filtering \cite{Plett2004,Plett2006,Baba2015} or adaptive observers \cite{Kim2010AObserver,Kim2012ComplementaryPrediction,Moura2013}. The choice of battery model is a trade-off between parsimony/computational resources, and accuracy/flexibility. Equivalent circuit models are ubiquitous and widely employed \cite{Plett2004,Plett2006}, but may suffer from a lack of accuracy across the wide range of operating points experienced in real world usage \cite{Gomez2011}. Higher fidelity `physics-based' models derived from porous electrode theory are also available \cite{Doyle1993,Fuller1994}, but are generally considered too complex and computationally demanding for state of health tracking despite recent improvements \cite{Jokar2016,Marquis2019,Sulzer2019a,Sulzer2019,Chu2019AElement}. 

In contrast, data-driven methods try to estimate state of health either directly from raw current, voltage and temperature measurements, or from features such as time spent in certain operating regions. Machine learning techniques used for this purpose include Gaussian process (GP) regression \cite{Richardson2019,Yang2018ACurve}, support vector machines \cite{Klass2014AOperation} and neural networks \cite{Chaoui2017AgingNetworks,Li2021OnlineNetworks}. Data-driven health estimation has mainly been investigated using laboratory data under controlled conditions at relatively small scales. Many publicly available experimental laboratory datasets exist \cite{dosReis2021EnergyIt}, but they are limited in size.

Supplementing laboratory aging data, field data from end users enables failure detection, gives insight into real-world performance, and improves understanding of manufacturing and usage variability \cite{Sulzer2021TheData}. There have been few studies of battery SoH estimation using field data, even though this presents a significant opportunity to broaden understanding of performance \cite{Song2020IntelligentAnalysis,Wang2020AData}. Although existing demonstrators show promise, their `black-box' nature reduces interpretability in comparison to model-driven approaches. Additionally, if charging and discharging patterns are highly dynamic, it might not be possible to calculate consistent features over the battery lifetime. So-called `hybrid' models that combine data-driven and physics-based methods \cite{Aykol2021PerspectiveCombiningLifetime} offer promise by balancing robustness, flexibility and transparency. 

\begin{figure}[t]
    \centering
    \includegraphics[width=\textwidth]{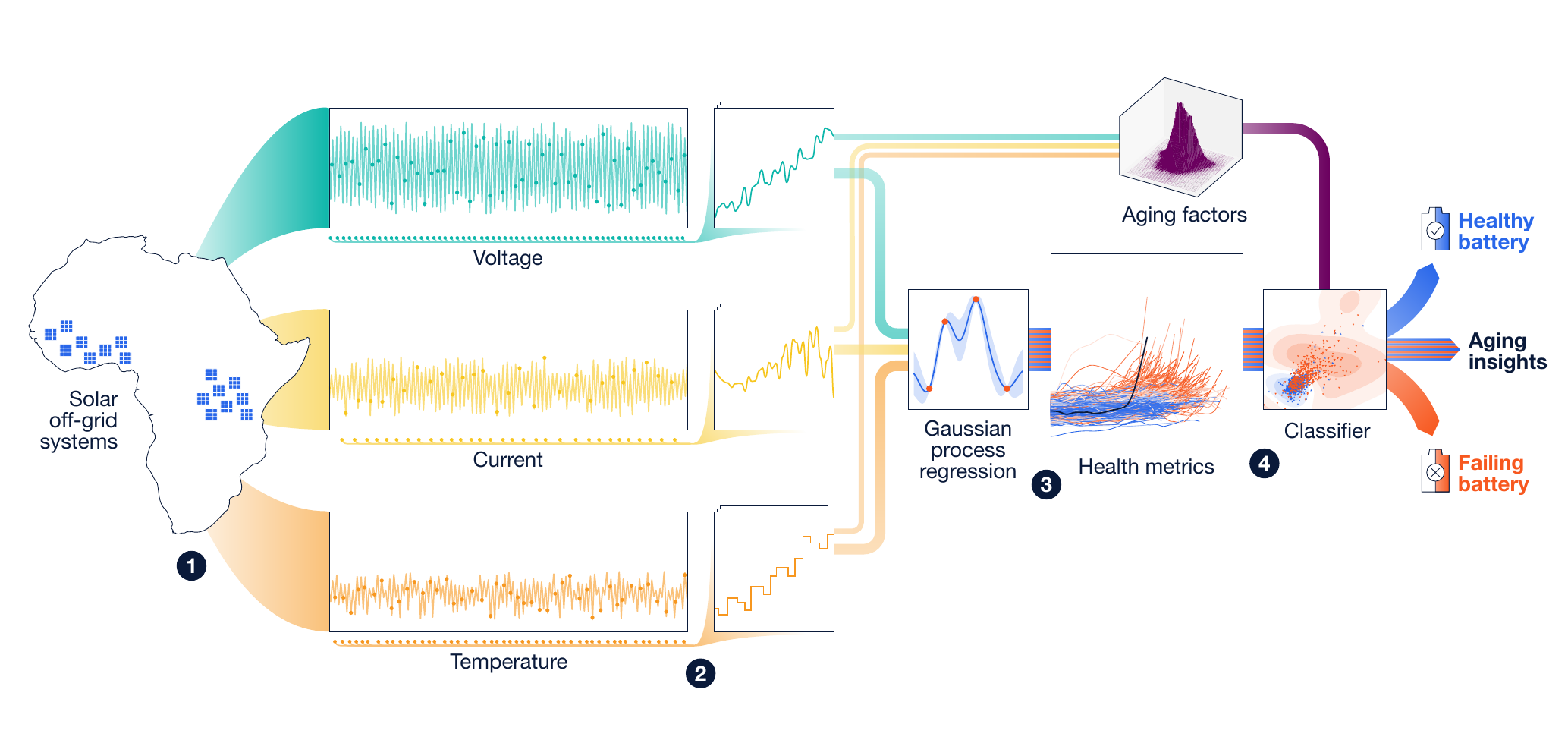}
    \caption{Workflow, left to right: (1) Data are retrieved from PV-connected batteries---the raw dataset contains 620 million rows of current, voltage and temperature measurements. (2) Data are sub-sampled for computational efficiency to yield 39 million rows from an average of 261 charging segments per battery. (3) GP regression is applied to obtain estimates for health metrics (calibrated internal resistance) over the lifetime of each battery. (4) Batteries are classified as healthy or failing up to 2 months prior to end of life, using these metrics plus other aging factors calculated from the raw data; this also gives insight into the key factors driving aging.}
    \label{fig:flow_chart}
\end{figure}

In real-world applications, constraints are imposed by changing usage conditions, less accurate sensors, lack of controlled usage and lack of prior knowledge of model parameters. Here, we implement physics-informed probabilistic machine learning for SoH estimation that is robust to changing operating conditions and data gaps. Our approach, illustrated in Fig.\ \ref{fig:flow_chart}, is demonstrated using measured current, voltage and temperature data from lead-acid batteries, but is applicable to any chemistry that can be represented with a low order electrical model, including lithium-ion cells. We employ Gaussian process regression \cite{Rasmussen2006,Sarkka2012a,Sarkka2013b} to construct health trajectories of 1027 batteries connected to photovoltaic systems in sub-Saharan Africa, using internal resistance as a health metric, since it may be estimated from data more accurately than capacity. Resistance is estimated as a function of instantaneous operating conditions, such as temperature and state of charge, enabling calibration of data, resulting in smooth estimates. Next, an end-of-life failure detection algorithm based on a Gaussian process classifier \cite{Rasmussen2006} predicts the probability of future failure by combining the estimated health trajectories with stress factors also extracted from the raw data. To train the classifier, batteries were labelled as failed or healthy using repair data from field workshops. We demonstrate, using 5-fold stratified cross-validation, that this technique gives 82\% balanced accuracy of end-of-life failure prediction at the time of failure versus a 66\% benchmark, and 73\% accuracy 8 weeks in advance of failure versus a 51\% benchmark. 

\section{Real-world dataset}

\begin{figure}[t]
    \centering
    \includegraphics[width=0.9\textwidth]{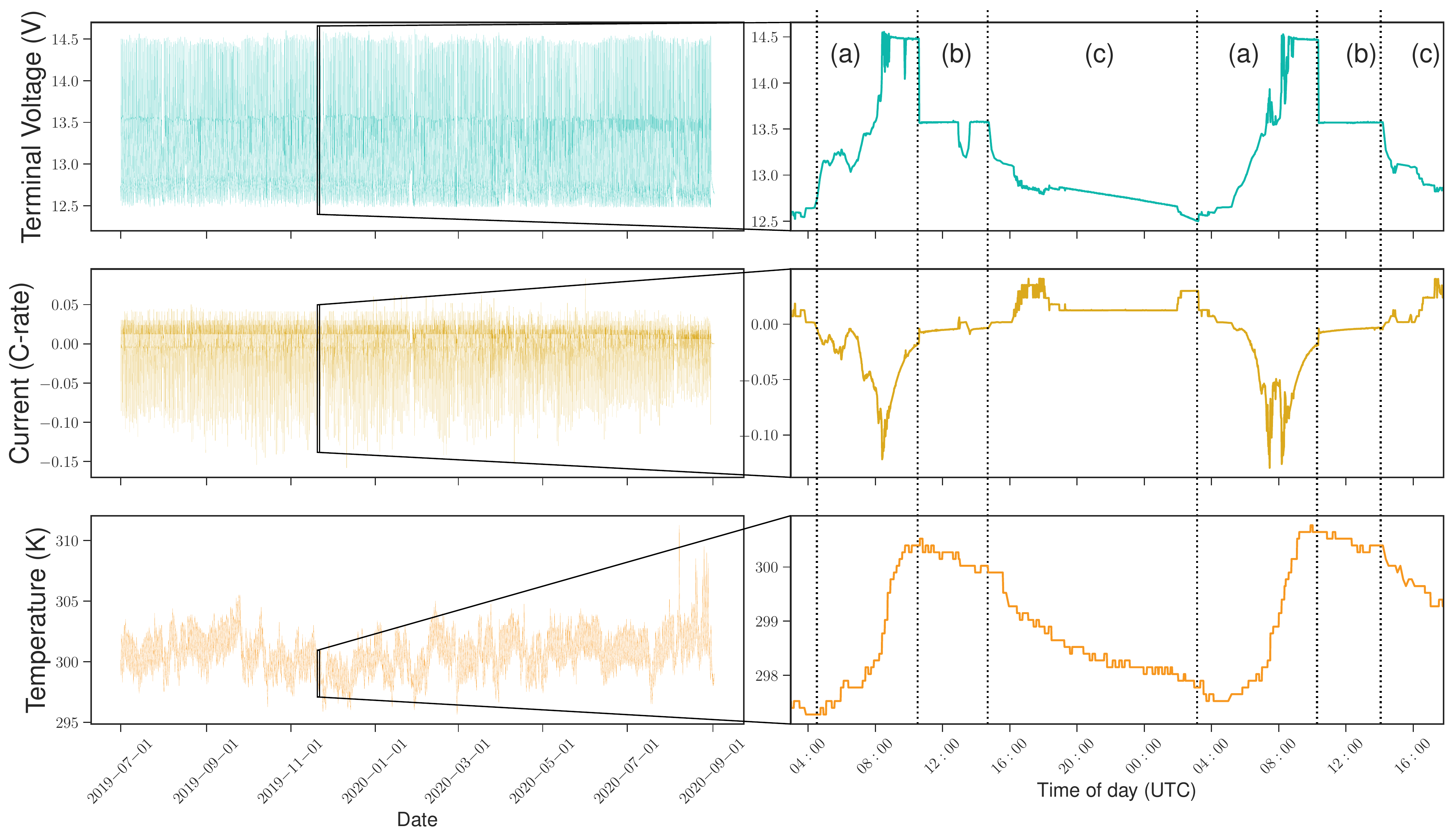}
    \caption{Solar off-grid systems exhibit variable temperatures and non-constant charging, with diurnal usage patterns including regions of (a) solar charge, (b) float charge, (c) discharge. Example data from a single battery system.}
    \label{fig:rw_data_1}
\end{figure}

Our dataset was generated by 1027 lead-acid batteries, each with nominal voltage 12 V (internally comprising 6 cells in series), nominal capacity 20~Ah, and attached to a 50 Wp photovoltaic panel. These systems are used for lighting, phone charging and small appliances, and are located across sub-Saharan Africa. Each battery was in use for 400-760 days, giving a total dataset size of 620 million rows (49 GB). This dataset is a small subset of the total number of systems deployed, and was selected to ensure that each time series was at least 400 days in length and also that the set contained approximately the same number of failed versus healthy batteries. A full explanation of the data selection process is given in Supplementary Material S1. Examples of the loading patterns experienced by the batteries in the dataset are given in Fig.\ \ref{fig:rw_data_1}.

There are some particular challenges for battery health diagnosis from the measured field data in this application. Firstly, the depth of discharge is commonly quite small, for example the majority of usage in our dataset is between 46--100\% SoC, making it difficult to observe changes in the discharge voltage curve caused by capacity changes as the cell ages. Secondly, since average currents are small (99th percentile approximately 0.1C, where C-rate is defined as the current divided by the nominal Ampere-hour capacity of the battery) and often constant, estimation of the internal resistance is numerically poorly conditioned, in other words small errors in measured voltage cause large errors in estimated resistance. These difficulties are compounded by unknown sensor accuracy. 

Furthermore, the average currents, temperatures and depths of discharge vary over time and across the population of batteries. We therefore require a methodology for health diagnosis that is robust to changes in usage and that does not require controlled diagnostic tests. In most battery systems, either capacity or internal resistance (or both) may be used as a health metric (e.g.\, \cite{Plett2011}, \cite{Remmlinger2011}). In lead-acid systems, internal resistance is a clear health indicator \cite{Kollmeyer2019AgingBatteries}, although the techniques outlined here can also be used to estimate capacity if sufficiently deep discharges are available. In our dataset, charging segments typically provide a more varied and higher amplitude input signal for resistance estimation compared with discharging segments, where the current is on average small and relatively constant. We therefore chose to estimate the resistance during charging as the state of health metric.

\section{Data-driven modelling}

We now describe the modelling and inference process, outlined in Fig.\ \ref{fig:flow_chart}, in detail. First, measured current, voltage and temperature data with non-uniform sampling were obtained from a central database, and suitable charging segments selected (see Methods)---on average, 261 such segments were identified per battery. Then, within each charging segment, data were interpolated onto a 1~minute uniform grid in time, and electrolyte acid concentration as a state of charge metric was estimated using Coulomb counting, with initial conditions calibrated by a known open circuit voltage versus state of charge relationship (see Methods). This process resulted in an average of 38000 rows of data for each battery, reduced 16-fold down from 603000 rows per battery.

The internal resistance of all batteries depends on age, current, temperature and state of charge. In lead-acid cells the reasons for this include non-linear kinetics \cite{Sulzer2019,Sulzer2019a}, nucleation and dissolution of lead sulfate \cite{Huck2020}, hydrolysis during charging \cite{Newman1997,DawnM.BernardiandMichaelK.Carpenter1995}, and degradation mechanisms such as sulfation, loss of active material and electrode corrosion \cite{Ruetschi2004,Schiffer2007}. Because of this, resistance estimates are not a reliable SoH metric unless they are first calibrated to remove the impacts of current, temperature and state of charge. Comprehensively modelling all the underlying physics would result in a model with a large number of parameters \cite{Newman1997}, and these are challenging to estimate from data \cite{Sulzer2019}. However, machine learning techniques can be used to learn from data the dependency of internal resistance on other factors. To ensure principled treatment of uncertainty associated with field data, we used a Bayesian approach, expressing the battery internal resistance as a Gaussian process over applied current, temperature, estimated electrolyte concentration (i.e.\ state of charge), and time.

\begin{figure}[t]
    \centering
    \includegraphics[width=0.8\textwidth]{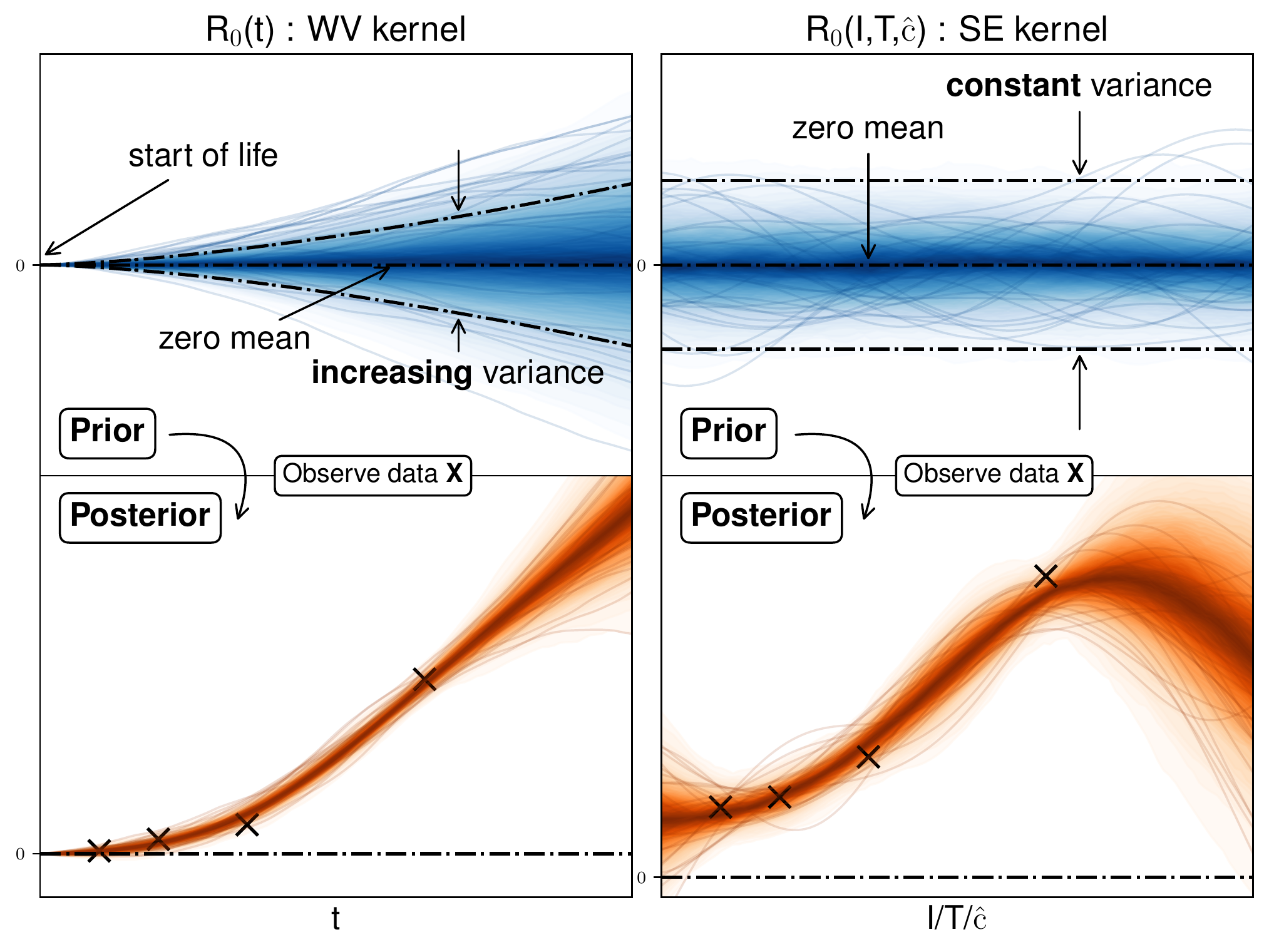}
    \caption{Illustration of the two kernels used to model battery internal resistance as a SoH metric. A Gaussian process defines a prior distribution over functions and is conditioned with a observations to produce a posterior distribution. Here, random draws from priors (top row) and posteriors (bottom row) are shown for a non-stationary Wiener velocity (WV) kernel used to model the time-dependency of resistance, and a stationary squared exponential (SE) kernel with short scale smoothness used to model the dependency of resistance on operating point.  Dotted lines show 1$\sigma$ in the priors. The WV kernel anchors the aging process to zero at start of life and extrapolates its last known trajectory, whereas the SE kernel is non-zero everywhere and reverts to the mean when extrapolated.
    }
    \label{fig:GP_priors}
\end{figure}

In contrast to parametric models \cite{Remmlinger2011}, Gaussian processes offer a flexible probabilistic technique that makes fewer assumptions about the structure of the underlying data \cite{Rasmussen2006}. We assumed that internal resistance over the lifetime of each battery consisted of a sum of two independent Gaussian processes, the first capturing the dependency on the instantaneous operating point, and the second the degradation. This improves computational efficiency, although it assumes that the dependence of resistance on age can be decoupled from the dependence of resistance on temperature, state of charge and current. We modelled the operating point dependency using a standard squared exponential kernel and the degradation dependency using a Wiener velocity kernel \cite{Solin2019a}, both shown in Fig.\ \ref{fig:GP_priors}. The former reflects our assumption that the variability of resistance with instantaneous temperature, state of charge and current should be relatively smooth. The latter is a non-stationary kernel \cite{ArnoSolin2016}, enabling degradation at the beginning of life to be zero for each individual battery, and then to grow as the battery ages. This means that extrapolation of future battery health, beyond available data, follows the trajectory learned from the data more accurately. %, instead of reverting to the prior mean. 

Gaussian process models have `hyperparameters' that describe the smoothness, magnitude and periodicity of the data being modelled. Fitting to data and then using the resulting model involves two steps. First, the hyperparameter values must be estimated, then the posterior distribution (i.e.\ mean and variance) of the internal resistance must be evaluated. Both of these steps can be computationally expensive, each by default scaling with $\mathcal{O}(n^3)$, where $n$ is the number of data points being fitted \cite{Rasmussen2006}, equal to approximately 38000 per battery. To overcome this challenge, we applied recursive techniques \cite{Sarkka2012a,Sarkka2013b} (see Methods). To benchmark our approach against an existing method \cite{Plett2004,Plett2006}, we  considered a separate case where $R_0$ follows a random walk through time and is assumed to be independent of operating conditions. This process is also controlled by a set of hyperparameters for which we found maximum-likelihood estimates using the same methodology.

\section{Constructing a consistent battery health metric}

To ensure that comparisons of health between  batteries were like-for-like, we estimated internal resistance as a function of temperature, state of charge, current and time in the down-selected dataset, and then chose a \emph{single} constant set of values of the operating conditions at which to evaluate SoH, for all batteries. This can be thought of as learning a function \mbox{$R_0 = f\left( T, I, \hat{c}, t \right)$} from data, using the GP technique described in the previous section, and then evaluating or slicing through this function at fixed values of all independent variables apart from time.

\begin{figure}
    \centering
    \begin{subfigure}{\textwidth}
        \includegraphics[width=0.99\textwidth]{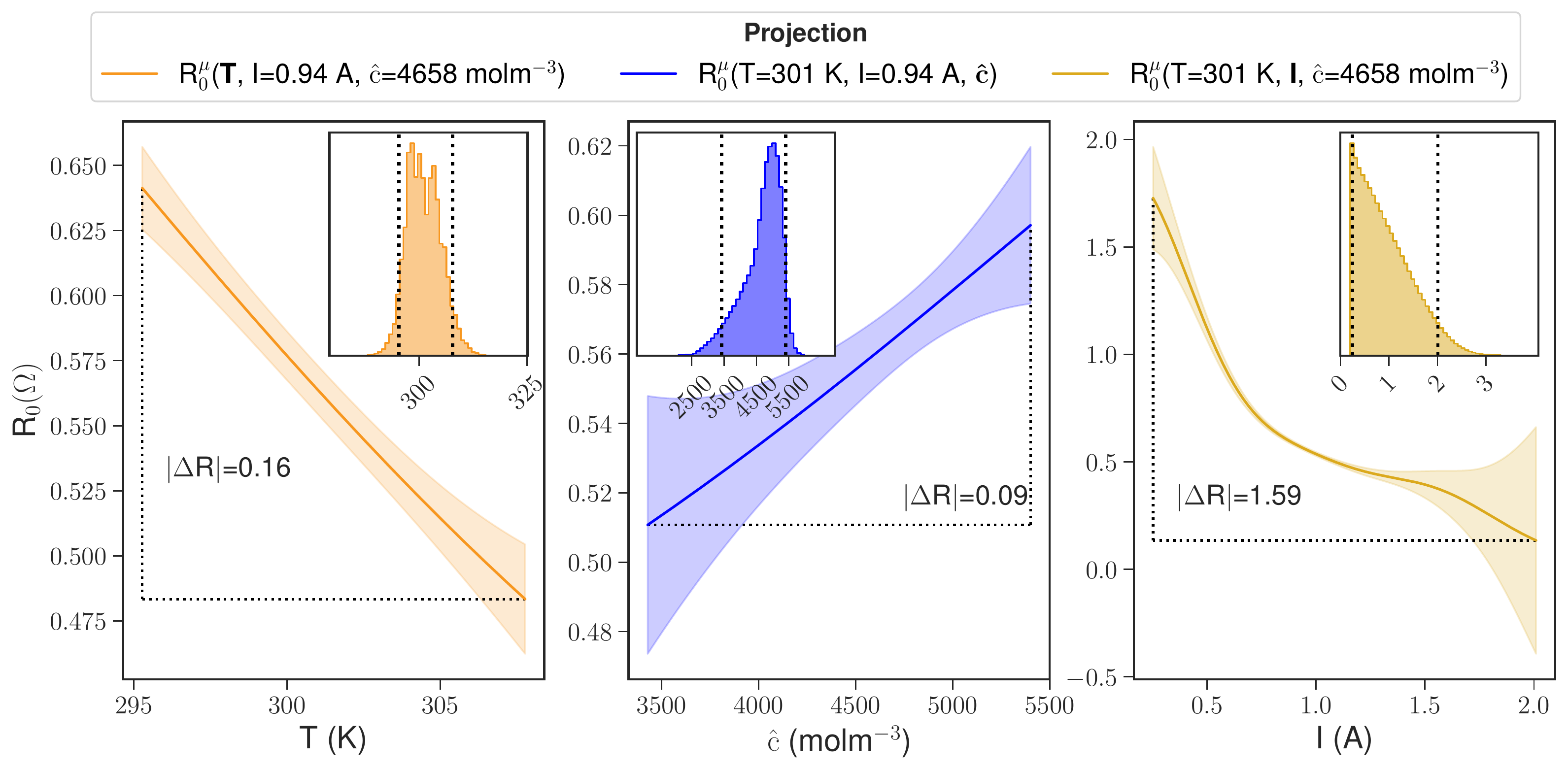}
        \caption{}
        \label{fig:op_dists}
    \end{subfigure}
    
    \begin{subfigure}{\textwidth}
        \includegraphics[width=0.99\textwidth]{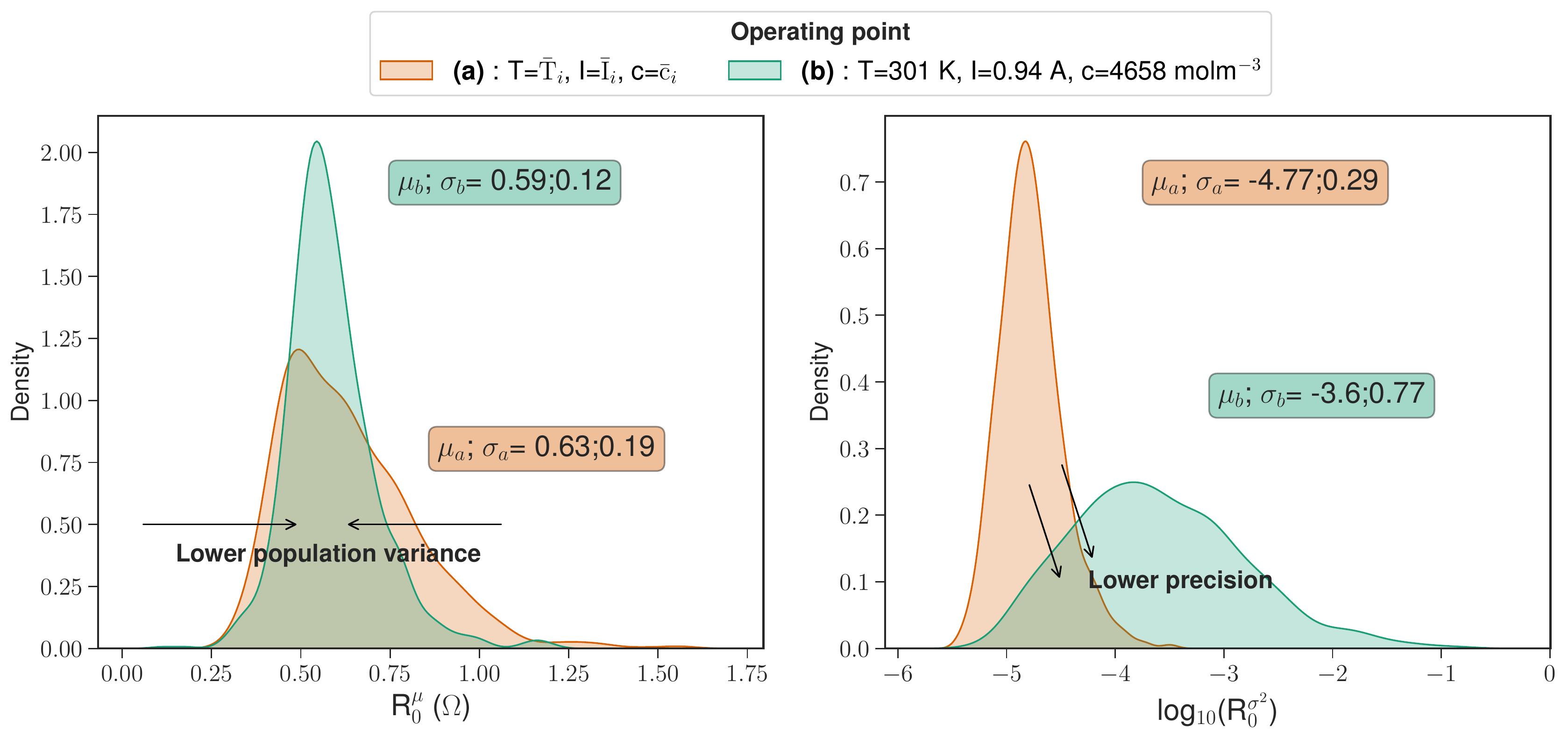}
        \caption{}
        \label{fig:common_op}
    \end{subfigure}
    \caption{(a) Internal resistance varies significantly with operating conditions, as shown by projections of estimated $R_0$ as a function of temperature, applied current and SoC (acid concentration) between 5th and 95th percentiles of each. Shaded regions show +/- 2$\sigma$ credible intervals. Insets show histograms of the independent variables for the down-selected dataset (Supplementary Material gives histograms for all data). (b) There is a trade-off between using a single population-wide calibration point vs.\ a local calibration point per battery. Left-hand image shows lower variance in estimated mean of $R_0$ values with a single calibration point. However, in this case the uncertainty associated with each $R_0$ estimate increases, as right-hand image shows.}
\end{figure}

Fig.\ \ref{fig:op_dists} shows the variability of internal resistance as a function of temperature, current and state of charge estimated at the population level (see Methods). A wide range of operating points are observed in the down-selected dataset, as shown in the Fig.\ \ref{fig:op_dists} insets, and consequently the estimated value of $R_0$ changes substantially with the instantaneous operating conditions. If this is not accounted for, then these effects would mask the variability caused solely by degradation. 

The function describing $R_0$ that is learned from data is consistent with battery physics. First, the inverse relationship between internal resistance and temperature is due to the Arrhenius dependence of reaction rate on temperature, and results in a variation of up to \SI{0.16}{\ohm} in estimated resistance. Second, the shape of internal resistance with respect to acid concentration is due to competing effects. The exchange current density is an increasing function of electrolyte acid concentration \cite{Sulzer2019a}, but the transport limitation caused by the reduced rate of dissolution of lead-sulfate during charging \cite{Huck2020} becomes dominant at higher states of charge. The size of this SoC effect was estimated to be approximately \SI{0.09}{\ohm} across our dataset, with relatively high uncertainty at lower SoC values, which are visited less often, as shown by the wider credible intervals. Finally, internal resistance reduces as current increases because the relationship between reaction rate and overpotential is nonlinear---usually described by Butler-Volmer kinetics \cite{Sulzer2019a}. Accounting for this dependency on applied current is key because it has the largest impact on $R_0$ in the observed operating range, resulting in changes of up to \SI{1.59}{\ohm}, although there is high uncertainty associated with the estimate at higher applied currents. This is caused by the batteries spending relatively little time at higher current, resulting in fewer data points and therefore a wider variance of the posterior predictive distribution obtained through GP regression. 

To calibrate estimates of battery internal resistance across all batteries we must choose a reference set of operating conditions (temperature, current, SoC). As indicated in Fig.\ \ref{fig:common_op}, there is a trade off between choosing a single fixed set of calibration conditions for the entire population of batteries versus using separate calibration conditions for each battery trajectory. To obtain a low variance estimate of the resistance of an individual battery, it is best to calibrate using the mean operating conditions for that individual battery, since conditions further from this are associated with higher uncertainty. However, a standardised population-wide calibration condition allows truly like-for-like comparisons to be made between batteries, and gives 37\% lower overall standard deviation, so this is what we chose. The remaining population variance reflects cell-to-cell variability at beginning of life resulting from manufacturing variations, storage time and conditions prior to field deployment.

\begin{figure}[t]
    \centering
    \includegraphics[width=0.8\textwidth]{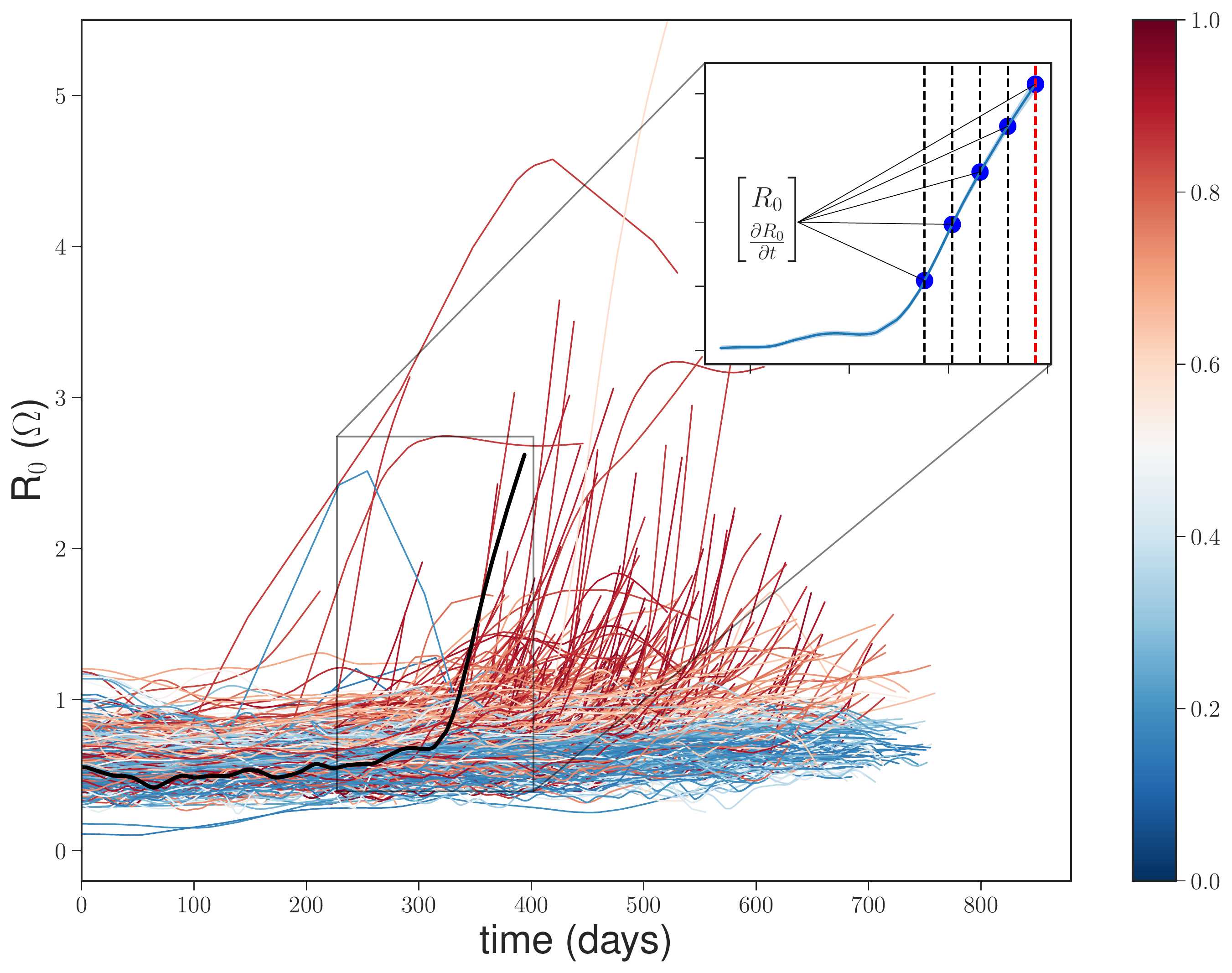}
    \caption{Our approach results in smooth SoH trajectories across the entire dataset of 1027 batteries, shown here as calibrated internal resistance profiles estimated from measured data. Color indicates the probability of a battery having failed at the end of each time line, as predicted by a classifier using the features calculated as inputs. The `knee point' is a key indicator of end of life, and as the inset shows, after this the resistance $R_0$  and its gradient $\partial R_0/\partial t$ increase.}
    \label{fig:feat_schem}
\end{figure}

As a result of the calibration process we  produce trajectories of $R_0$ that are only a function of time, all referred to a common operating point which was the population mean of temperature, current and acid concentration. The  trajectories are shown in Fig.\ \ref{fig:feat_schem}. In addition to the absolute value of resistance $R_0$, the derivative with respect to time $\partial R_0/\partial t$ is also a useful health indicator (calculation given in Supplementary Material). Together these two metrics capture whether the battery degradation is beyond the so-called 'knee point', i.e.\ the onset of accelerated degradation towards end of life. 

\section{Predicting and understanding battery end of life}
Having constructed consistent battery health trajectories from raw measured data by estimating internal resistance and calibrating it for operating conditions, we can now (i) validate the health metrics by using them to predict end-of-life failure, and (ii) gain insight into the factors that drive battery aging. To predict battery end of life due to aging, we used repair data provided by BBOXX Ltd.\ to label each battery as failed or healthy at the end its respective data time series, and trained a GP classifier \cite{Rasmussen2006} to predict this based on health and stress factors at various different prediction horizons. This provides an indirect method of validating the health estimates, although there is uncertainty associated with the labeling of healthy and failed batteries, i.e.\ there could be false positives and false negatives since repairs are driven by customer decisions. The available data were split into training and test sets, and the classification performance was measured by predicting whether a repair would occur in the test set.

We assessed the performance of this end-of-life classifier under multiple scenarios. First, we used only the health features $R_0$ and $\partial R_0/ \partial t$ described above as inputs. Second, to benchmark the performance, we used a health metric consisting solely of internal resistance fitted with a random walk, as is common in the literature \cite{Plett2004,Plett2006}, without consideration of operating-condition dependency. Third, we considered the case where the health metrics $R_0$ and $\partial R_0/ \partial t$ were augmented with stress factors known to affect lead-acid battery health \cite{Ruetschi2004}. The latter are features indicative of usage that are also extracted from the raw data, and they consisted of calendar age, charge throughput, cycle count, cumulative time spent at float charge voltage, as well as average temperature and voltage (see Supplementary Material S5 for calculation details). Fourth, we tested the performance using solely the stress factors, omitting information relating to the current state of health as indicated by resistance. To demonstrate the ability to predict end of life in advance with different horizons from zero to eight weeks, and with varying ratios of failed versus healthy batteries, we structured the classification performance tests with nested cross validation (see Methods). 

\begin{figure}[t]
    \centering
    \begin{subfigure}{0.48\textwidth}
        \includegraphics[width=0.95\textwidth]{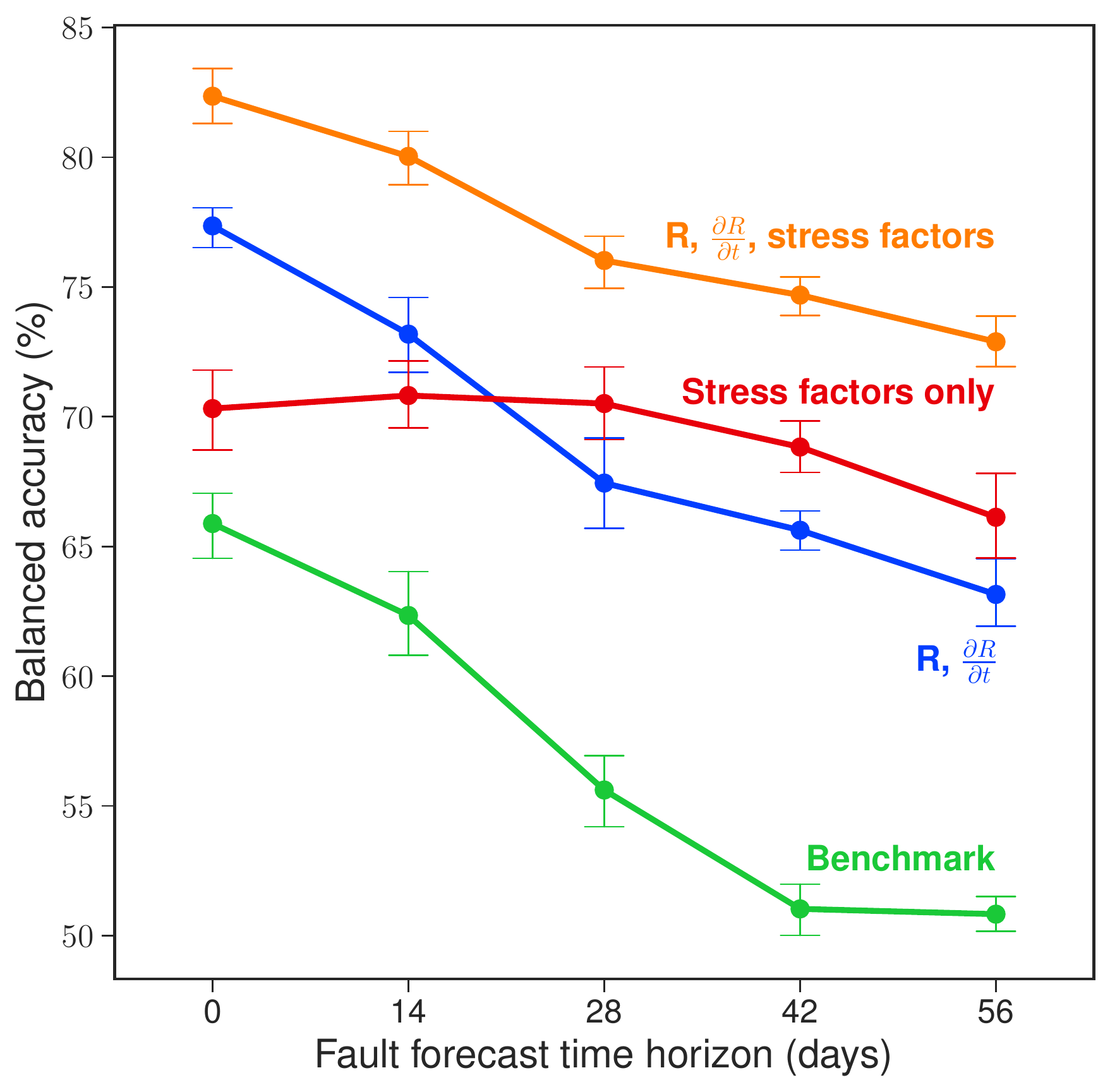}
        \caption{}
    \end{subfigure}
    \begin{subfigure}{0.49\textwidth}
        \includegraphics[width=.97\textwidth]{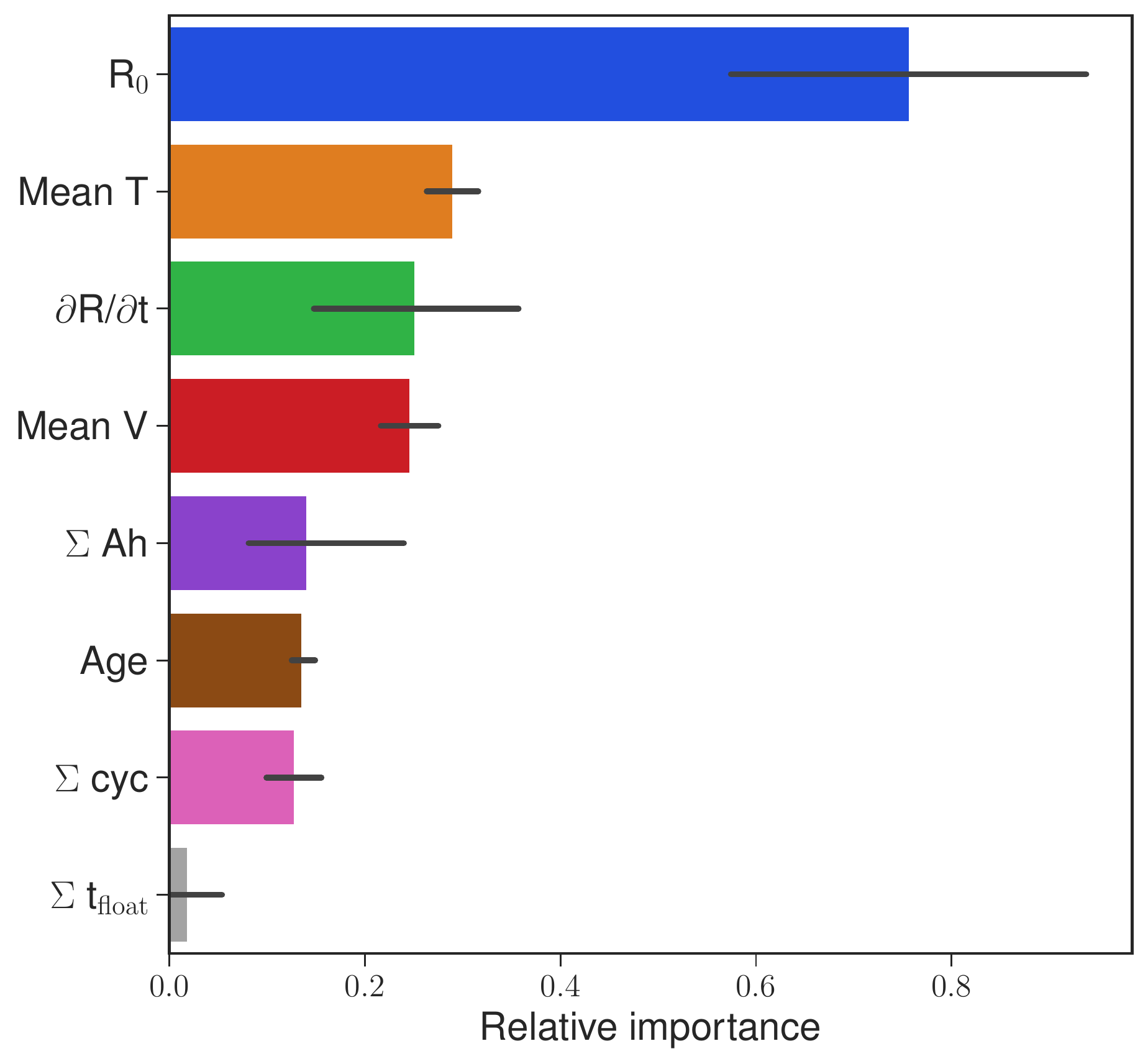}
        \caption{}
    \end{subfigure}
    
    \caption{(a) Combining estimated SoH (resistance and gradient of resistance) with aging stress factors gives best overall end-of-life prediction. Classification performance is quantified using balanced accuracy, for prediction horizons from 0-56 days. Error bars show the standard deviation in the average test accuracy grouped by test set, including a varying proportion of failed batteries. (b) Internal resistance, mean temperature, rate of change of internal resistance with time and mean voltage are important predictors of end of life. The relative importance of various inputs was quantified using the inverse length scales of the hyperparameters averaged across training sets. The standard deviation of each inverse length scale is indicated by the error bar.}
    \label{fig:perf_fig}
\end{figure}

The classifier performance as a function of prediction horizon and according to the various scenarios is shown in Fig.\ \ref{fig:perf_fig}(a). The best overall end-of-life diagnosis and prediction performance occurs when we combine both the current state of health, as indicated by estimated resistance and its gradient, with the stress factors. In this case 82\% balanced accuracy (see equation \ref{eqn:bal_acc} in Methods for definition) is achieved at end of life and 73\% when the time horizon for prediction is extended to 8 weeks. The Supplementary Material gives the full confusion matrix showing classifier performance. In comparison, the benchmark approach and the approach using only estimated SoH---without stress factors---give a lower accuracy which rapidly drops with increasing prediction horizon. This is due to the absence of knee points in SoH as one moves further back in time from the end of life; two months before they fail, relatively few batteries have experienced the onset of accelerated degradation. At this point the benchmark model performance is near 50\% balanced accuracy, which is equivalent to random classification. The classifier using only the stress factors---without resistance estimates---performs well in comparison to the benchmark, achieving an average accuracy of 68\% over all horizons. This indicates that aging is being driven by usage conditions and calendar life, and is fairly consistent across the population, although introducing battery-specific SoH information near end of life significantly improves  predictive accuracy.

The classification technique also gives insight into the factors driving aging. Relative importance was quantified using the inverses of the length-scale hyperparameters of the classifier. Resistance and its rate of change over time are important end-of-life indicators, but other factors, in particular mean temperature and mean voltage, are significant, as shown in Fig.\ \ref{fig:perf_fig}(b). Temperature is known to have a complex impact on lead-acid battery degradation, and the full dataset spans more than \SI{20}{\celsius} range (see Supplementary Material). Elevated temperatures, especially during charging, may improve lifetime due to improved solubility of lead sulfate \cite{Ruetschi2004}, although they also increase electrode grid corrosion. Similarly, mean voltage is indicative of mechanisms such as sulfation, which occurs at low voltage \cite{Ruetschi2004}. Although some aging factors, such as cycle count, are considered unimportant, this is likely because they are correlated with other factors such as charge throughput and age (a correlation matrix is given in Supplementary Material S5).

\section{Conclusions}
Real-world operating data from battery systems in the field may be used to detect end-of-life failure before it happens, improving maintenance, value, safety and customer experience and giving insights into degradation and performance. We have developed and demonstrated at large scale a data-driven approach for battery state of health estimation and end-of-life prediction using only measured current, voltage and temperature data whilst in use, without requiring controlled cycling or systems to be taken offline. Using data from 1027 solar off-grid lead-acid batteries, each running for 400-760 days, we obtain an end-of-life prediction performance of 73\% balanced accuracy, 8 weeks prior to end of life, rising to 82\% accuracy at end of life. This accuracy was achieved by combining estimates of state of health with aging stress factors also calculated from the measured data. We used probabilistic machine learning techniques to learn internal resistance as a function of current, temperature, state of charge and time, enabling us to calibrate state of health to consistent reference conditions across the entire population dataset. The success of the approach may be explained by the combination of a population-wide health model with a battery-specific indicator that becomes increasingly important towards end of life. These techniques are applicable to any battery that can be represented with a simple electrical circuit model. Broadly, this work highlights the opportunity to complement laboratory battery data with large field datasets analyzed through machine learning to improve performance and understanding.
% ------------------------------------------------------------------------------------------------
% ------------------------------------------------------------------------------------------------

\section{Methods}

\subsection{Data selection}

To obtain a clean dataset and reduce the computational load required for model training, several pre-processing steps were necessary. First, segments of charging data were chosen over the lifetime of each battery. To be eligible, segments had to meet the conditions listed in Table \ref{tab:seg_conds}.
\begin{table}[H]
    \centering
    
    \begin{tabular}{rr}
    \toprule
    Charging segment duration & $>$ \SI{6000}{\second} \\
     Starting voltage range & \SI{11.5}{\volt}-\SI{12.9}{\volt}   \\
     Starting current & $<$ \SI{0.1}{\ampere} \\
     max(Voltage in segment) & $>$ \SI{14}{\volt} \\
     max(Time gap in recorded data) & $<$ \SI{610}{\second} \\
     \bottomrule
    \end{tabular}
    
    \caption{Qualifying charging segment conditions}
    \label{tab:seg_conds}
\end{table}

These conditions ensured that charging segments covered a reasonable state of charge range. Additionally, each charging segment was truncated to include voltages only up to \SI{14}{V} due to increased uncertainty in estimating state of charge at higher voltages when using the method described in section \ref{subsection:soc}, because the magnitude of the side reactions increases exponentially with terminal voltage. After down-selecting appropriate charging segments, data were interpolated to a 1-minute time grid using piece-wise cubic hermite interpolation \cite{Carlson1980}. Data points where $I\ts{t}<$\SI{0.2}{A} or $V\ts{OC,t}<V_t$ were removed to improve conditioning and to make sure the open circuit voltage was consistent.

\subsection{Inferring acid concentration from current and voltage data}
\label{subsection:soc}
In lead-acid batteries, electrolyte acid concentration is a direct measure of state of charge---the latter is the normalised version of the former, but requires the maximum concentration to be known, and this changes with age. We therefore simply use acid concentration as a state of charge indicator throughout, and use the two terms interchangeably in this paper. To infer acid concentration from measured data, we first measured the battery open circuit voltage in a laboratory (see Supplementary Material S2), then used this to determine initial acid concentration from measured voltage at points where the charging current was at a minimum. The required open circuit voltage curve was measured experimentally using the galvanostatic intermittent titration technique (Biologic SP-150 potentiostat), placing the battery in a thermal chamber (Binder MK53) at \SI{25}{\celsius}. From this data, the electrolyte volume was also inferred by a least squares fit using Bode's well known result \cite{Bode1977}. Finally, the trajectory of acid concentration for each charging segment was obtained by Coulomb counting from the initial concentration and using the measured current data, accounting for the known side reactions in lead-acid systems \cite{Newman1997,DawnM.BernardiandMichaelK.Carpenter1995} by using a lumped term for the gassing reactions \cite{Schiffer2007} according to
\begin{equation}
    \frac{d\hat{c}\ts{t}}{dt} = \frac{I\ts{t}-I\ts{gas,0}e^{c_T(T\ts{t}-T\ts{0})+c_V(V\ts{t}-V\ts{0})}}{FV\ts{elec}},
    \label{eqn:Coulomb_counting}
\end{equation}
where $I\ts{t},T\ts{t},V\ts{t}$ are the measured current, temperature and terminal voltage, $F$ is Faraday constant, $V\ts{elec}$ is the estimated electrolyte volume, and gassing current parameters $I\ts{gas,0},c_T,T_0,c_V,V_0$ were from literature \cite{Schiffer2007}. The substantial uncertainty due to the variation in these parameters over the lifetime of the battery was taken into consideration by projecting the input uncertainty in the acid concentration $\hat{c}$ to be measurement noise variance in GP regression (equations \ref{eqn:var_accum}) . 

\subsection{Input data normalisation}
To ensure comparability of hyperparameters, the current, temperature and estimated acid concentration $I\ts{t}, T\ts{t},\hat{c}\ts{t}$ were normalised using the population-level moments according to
\begin{equation}
    X = \frac{x-\bar{x}}{\sigma\ts{x}} ~, ~ x\in \{T\ts{t}, I\ts{t}, \hat{c}\ts{t} \},
    \label{eqn:normalize}
\end{equation}
where $\bar{x}$ and $\sigma\ts{x}$ represent the population mean and variance of the down-selected dataset. A normalised time-scale was obtained by dividing the time since beginning of life in seconds by \SI{34560000}{\second}, i.e.\ a nominal 400 days life expressed in seconds, bringing it to a similar range as the other inputs. This method of normalisation (\ref{eqn:normalize}) was also used for the inputs for the GP classifier.

\subsection{Modelling internal resistance as a Gaussian Process}

\subsubsection{Battery-specific estimates} \label{b_wise_R0}

Given that the C-rates during charging are low ($<$0.2C), we may ignore the effect of concentration overpotentials \cite{Sulzer2019,Sulzer2019a}. Therefore the terminal voltage is given by the sum of the open circuit voltage and a lumped linearised internal resistance term,
\begin{equation}
    V\ts{t} = V\ts{0}(\hat{c}\ts{t}) + R\ts{0}(t, I\ts{t},T\ts{t},\hat{c}\ts{t})I\ts{t} + \epsilon\ts{t} ~,~~ \epsilon\ts{t} \sim N(0,\sigma^2\ts{n,t})
    \label{eqn:regression}
\end{equation}
where $V\ts{t}$ is the terminal voltage and $t, I\ts{t}$, $T\ts{t}$, $\hat{c}\ts{t}$ are the (normalised) time since beginning of life, applied current, measured temperature and estimated bulk sulfuric acid concentration at time t respectively. The experimentally obtained open circuit voltage as a function of acid concentration is $V_0$ and the dependency $R_0(I\ts{t},T\ts{t},\hat{c}\ts{t},t)$ is modelled by a zero mean Gaussian process, 
\begin{equation}
    R_0(x) \sim \mathcal{GP}(0,k(x)) ~,~ x \in \{t,I,T,\hat{c}\ts{t}\},
    \label{eqn:GP_def}
\end{equation}
where $k$ is the covariance function. We chose to model $R_0$ as a sum of two GPs, where the degradation process as a function of time is described by a Wiener velocity (WV) kernel \cite{ArnoSolin2016} and the dependency of $R_0$ on operating point  $(I\ts{t},T\ts{t},\hat{c}\ts{t})$ is described by a squared exponential (SE) kernel, such that
\begin{equation}
    k(I,T,\hat{c},t) = \sigma^2\ts{f,0} \left(\frac{\min^3(t,t')}{3}+|t-t'|\frac{\min^2(t,t')}{2} \right) + 
    \sigma^2\ts{f,1} \exp\left(\sum_{x\in \{I,T,\hat{c}\}}-\frac{(x-x')^2}{2l\ts{x}^2}\right),
    \label{eqn:kernel}
\end{equation}
where $|.|$ denotes the absolute distance between two points. The choice of the WV process, used in target tracking applications \cite{Bar-ShalomYaakov2001Ewat}, allows for better extrapolation outside the observed data because the kernel is non-stationary. In comparison, using the zero-mean SE kernel would result in extrapolation over longer time horizons tending to the prior distribution. Examples of random draws from the prior distributions for each kernel are shown in Fig.\ \ref{fig:GP_priors}.

Extrapolation over time is necessary in our case because the down-selected charging segments are not evenly distributed over time across the population. Extrapolation is thus required to give estimates of $R_0$ at points 0-56 days preceding the data series end for each battery. Additionally, expressing $R_0$ as a sum of kernels, equation \ref{eqn:kernel}, makes the assumption that the degradation is purely additive, and independent of the operating point chosen. This significant simplification reduces the degrees of freedom of the system in comparison to a product of kernels over all inputs and makes inference computationally lower cost.

To fit the model defined by equations \ref{eqn:regression}-\ref{eqn:kernel}, the hyperparameters to be estimated are the two process variances, $\sigma^2\ts{f,0},\sigma^2\ts{f,1}$, and length scales $l_x$ across the inputs, $(I\ts{t},T\ts{t},\hat{c}\ts{t})$. Special treatment was given to the noise $\epsilon_t$ in equation \ref{eqn:regression}. There may be considerable uncertainty in the open circuit voltage function $V_0(\hat{c})$, for example due to hysteresis, as well as in the estimate of acid concentration $\hat{c}$, due to uncertainty in the parameters in equation \ref{eqn:Coulomb_counting}. To account for this, we assumed a 10\% uncertainty in $d\hat{c}/dt$, together with a \SI{150}{\milli\volt} standard deviation caused by voltage measurement and open circuit voltage uncertainty, giving a total variance per charge segment as
\begin{subequations}
\begin{align}
\sigma^2\ts{n,t} & = 0.0225 + \text{Var}(\hat{c}\ts{t})\left( \left.\frac{d V\ts{0}}{d\hat{c}}\right|_{\hat{c}}\right)^2 \\
\text{Var}(\hat{c}\ts{t}) &= \sum_{t}{0.01 \Delta \hat{c}\ts{t}^2},
\end{align}
\label{eqn:var_accum}
\end{subequations}
where $\Delta \hat{c}\ts{t} = \left( dc\ts{t}/dt \right) \Delta t$, assuming $dc\ts{t}/dt$ is constant in each time segment $\Delta t$. In summary, the GP regression includes heteroskedastic noise $\sigma\ts{n,t}^2$ which we pre-calculate rather than estimate as a hyperparameter in the fitting process.

To impose smoothness on the function $R_0$ over all input dimensions, we assumed a prior distribution over hyperparameters,
\begin{equation}
    p(\sigma\ts{f,0},\sigma\ts{f,1},l\ts{T},l\ts{I},l\ts{$\mathrm{\hat{c}}$}) = \prod_{m \in \{0,1\}} \mathcal{\chi}(\sigma\ts{f,m},\text{k}=1,\text{s}=0.2) \prod_{x \in \{I,T,\hat{c}\}}{\Gamma^{-1}\left(l\ts{x},\mathrm{\alpha=1},\mathrm{\beta}=2\right)},
    \label{eqn:hyperpriors}
\end{equation}
where $\mathcal{\chi}$ represents the chi distribution and $\Gamma^{-1}$ the inverse gamma distribution. The chi distribution with the degree of freedom $k=1$ is equal to a half-normal distribution with standard deviation $s=0.2$. We assumed an inverse gamma prior for the length scales, with shape and scale parameters $\alpha$ and $\beta$ chosen to give a mode of 1.

To estimate the posterior distribution of $R_0$ for each battery and to recover maximum-a-posteriori (MAP) estimates of the hyperparameters, we employed a recursive estimation framework for GPs \cite{Hartikainen2010,Sarkka2012,Sarkka2013b}, where they are interpreted to be the solution of a stochastic partial differential equation of the form 
\begin{equation}
    \frac{\partial R_0(t,I,T,\hat{c})}{\partial t} = F R_0(t,I,T,\hat{c}) + Lw(t,I,T,\hat{c}),
\label{eqn:GP_SDE}
\end{equation}
where the transition matrix $F$, dispersion matrix $L$ and the properties of the white noise process $w(t,I,T,\hat{c})$ are determined by the kernel function, equation \ref{eqn:kernel}.
This framework allows the use of standard Kalman filtering and smoothing techniques \cite{ArnoSolin2016,Sarkka2013} to estimate both the posterior distribution of $R_0$ and the so-called energy function, that is the negative unnormalised logarithm of the posterior probability of the hyperparameter vector. Crucially, this method scales as $\mathcal{O}(n)$ over the number of data rows, which in our case is order 10$^4$ for each battery.

To obtain a finite-dimensional representation of the system in equation \ref{eqn:GP_SDE}, we used a similar approach to that of S\"arkk\"a et al.\ \cite{Sarkka2012}. First, we applied a k-means algorithm to choose 20 representative points across $(I\ts{t},T\ts{t},\hat{c}\ts{t})$, at which to estimate $R_0$ through the lifetime of the battery. In order to obtain estimates for $R_0$ for the observed $(I\ts{t},T\ts{t},\hat{c}\ts{t})$ through all charging segments, the predictive distribution of the GP over the operating points was calculated and added to the value predicted by the degradation GP.

Using these recursive techniques, we fitted regression hyperparameters across the entire population of batteries, consisting of 39 million rows of data, in approximately 80 minutes using an Apache Spark\textsuperscript{TM} cluster running 30 cores.

\subsubsection{Population level model over operating points}
Population level estimates of $R_0$ as a function of all the operating points were calculated by fitting a Gaussian process over the individual battery $R_0$ estimates (i.e.\ the solutions of equation \ref{eqn:GP_SDE}) at each battery's mean operating point at the beginning of life. This was done using the standard batch GP approach \cite{Rasmussen2006}, with the SE kernel and hyperprior for the magnitude and lengthscales equivalent to the recursive case. The hyperprior including the measurement noise was then
\begin{equation}
     p(\sigma\ts{f},\sigma\ts{n},l\ts{T},l\ts{I},l\ts{$\mathrm{\hat{c}}$}) = \mathcal{\chi}(\sigma\ts{n},\text{k}=1,\text{s}=0.1)  \mathcal{\chi}(\sigma\ts{f},\text{k}=1,\text{s}=0.2) \prod_{x \in \{I,T,\hat{c}\}}{\Gamma^{-1}\left(l\ts{x},\mathrm{\alpha=1},\mathrm{\beta}=2\right)}.
\label{eqn:GP_batch_hyper_prior}
\end{equation}
In addition, we added the variance estimates for each $R_0$ values to the measurement noise. Given MAP estimates retrieved by the L-BFGS-B algorithm \cite{doi:10.1137/0916069}, the posterior means and variances for the population level $R\ts{0}$ function were given by
\begin{subequations}
\begin{align}
    \mu\ts{p} & = k\left(x^{*},X\right)\left[k(X,X) + \sigma\ts{n}^2\mathbf{I} + \sigma^2\ts{R}(X)\right]^{-1}R\ts{0}(X) \\
    \sigma^2\ts{p} & = k(x^{*},x^{*}) - k(x^{*},X)\left[k(X,X) + \sigma\ts{n}^2\mathbf{I} + \sigma^2\ts{R}(X)\right]^{-1}k(X,x^{*}),
\end{align}
\end{subequations}
where $k$ is the SE kernel function and $R\ts{0}(X)$ and $\sigma^2\ts{R}(X)$ denote the mean and variance of the battery-wise $R_0$ estimates at operating point $X$ --- in this case $X$ was the mean operating for the battery. The predictive distribution was calculated on a grid $x^{*}$ such that the range for each input variable (temperature, applied current and estimated acid concentration) was varied between its 5th and 95th percentile in turn while keeping the other two constant at the population mean. 

\subsubsection{Benchmark model}

To benchmark our approach, we used the same model (equation \ref{eqn:regression}) but without the dependency of $R_0$ on the operating point. Furthermore, we assumed that $R_0$ followed a random walk through time, an approach commonly taken in the literature to adapt parameters to data \cite{GregoryL.Plett2015}. Using this recursive approach, the tuning parameters are the process and noise covariances as well as the initial variance estimate for $R_0$, which are often set manually. For consistency with the main approach, we estimated the process noise covariance using maximum likelihood and set the initial variance to be a constant multiple of the process covariance and the noise covariance was calculated in the same way (equations \ref{eqn:var_accum}) as for the main method.

\subsection{End-of-life prediction using Gaussian process classification}

To classify the batteries as failed or healthy at various time horizons up to the end date of each battery data series, we used a standard GP classification framework \cite{Rasmussen2006}. The inputs $R_0, \frac{\partial R_0}{\partial t}$ were extrapolated as required from the last observed charging segment to the appropriate point in time preceding the end of the time series. In all cases,  inputs were fed into a GP classifier using a SE covariance function with automatic relevance detection,
\begin{equation}
    k\ts{class}(x,x') = \sigma\ts{f}^2 \exp\left(\sum_x -\frac{(x-x')^2}{2l_x^2}\right)
    \label{eqn:gpclass}
\end{equation}
with an uniform hyperprior. Maximum likelihood estimates for the hyperparameters were then obtained by minimizing the negative log marginal likelihood of the data, using the GP classifier implementation in the Scikit-learn toolbox \cite{scikit-learn}. The performance metrics chosen for the classifer were such that the unevenness in labelling (536 healthy, 491 failed) was taken into account. The balanced accuracy metric was calculated as the average of classifier sensitivity and specifity,  
\begin{equation}
    \text{Balanced accuracy} = \frac{1}{2}(\text{Sensitivity+ Specifity}) = \frac{1}{2}\left(\frac{\text{TP}}{\text{TP+FN}}+\frac{\text{TN}}{\text{TN+FP}}\right),
    \label{eqn:bal_acc}
\end{equation}
where TP, TN, FP, FN represent the true positive, true negative, false positive and false negative rates respectively.

To quantify classifer performance, first we used 5-fold stratified cross validation to split the dataset into training and test sets. The training set in each case had a 48/52\% proportion of failed vs.\ healthy batteries using data up to the end of each time series, training 20 classifiers altogether (4 sets of inputs, and 5 training sets). Then, for each outer cross-validation stage, test subsets were selected, each containing varying proportions of failed batteries. These subsets were chosen by keeping the healthy battery set fixed and then randomly sampling 40\%, 60\% and 80\% of the failed batteries in the test set, repeated ten times, plus the case when 100\% of the failed batteries was used, giving a total of 3100 test cases. All performance metrics reported are the average for the test sets using 5-fold cross validation for each test case.

\section{Acknowledgements}
The authors thank BBOXX Ltd.\ for access to data from PV-connected battery energy storage systems; D.\ Lovell for assistance with Fig.\ \ref{fig:flow_chart}; V.\ Sulzer, R.\ Drummond, J.\ Reniers and S.\ Cooper for manuscript feedback. We acknowledge funding from the Faraday Institution (EP/S003053/1, grant number FIRG003) and Shell Foundation (agreement 22077).

\section{Author contributions}
Conceptualization, D.A.H.; data curation, software, visualization, A.A.; writing -- original draft, review, editing, A.A. and D.A.H.; supervision, project administration, funding acquisition, D.A.H. 

\section{Competing interests}
The authors have filed a patent related to this work: GB Application No.\ 2105995.1, dated 27 April 2021. D.A.H. is co-founder of Brill Power Ltd., and is a technical advisor at Habitat Energy Ltd. A.A. declares no competing interests.

\printbibliography



\section*{Supplementary material}

# Supplementary Material for 'Predicting battery end of life from solar off-grid system field data using machine learning'

Antti Aitio, David Howey

July 29, 2021

## 1 Selection of battery data

The database maintained by off-grid solar provider BBOXX Ltd. contains time series data for over 300,000 batteries. The data include different chemistries, cell capacities and manufacturers. Ensuring comparability of batteries was crucial, so several steps were taken to obtain a consistent dataset. We obtained the final set of 1027 batteries through the following steps:

1. We chose lead-acid batteries from single manufacturer with nominal capacity 20 Ah.

2. We defined the lifetime of a battery as the time since the 'activation date'. We chose batteries with a lifetime of over 400 days, with an activation date after 2018-09-01.

3. For those batteries that entered repair at some point after activation, end of life was defined as the date they entered repair. The dataset was filtered down so only batteries that either had no repairs or entered repair due to loss of capacity (as defined by BBOXX) were included. The (initial) failed set consisted of all the batteries that had been diagnosed with a loss of capacity when taken into a repair shop. To make up the total dataset with an approximate 52/48 split between healthy and failed cells, we used stratified sampling with respect to battery lifetime to generate a set of healthy batteries from the overall population.

4. We removed any batteries that had large gaps in their telemetry data (over 30 days).

5. We removed any batteries that had tampering alerts during operation.

6. We filtered out telemetry data for each battery where temperature readings were clearly incorrect.

7. We removed any batteries that did not have enough charging segments. We calculated the number of charging segments available for each battery as described in the Methods section and filtered out batteries with fewer than 50 qualifying segments. Additionally, batteries were omitted where the last qualifying segment was more than 30 days before the end of the dataset series.

8. Batteries that entered repair were labelled 'faulty', and their timeseries were truncated to only include data up to the repair date.

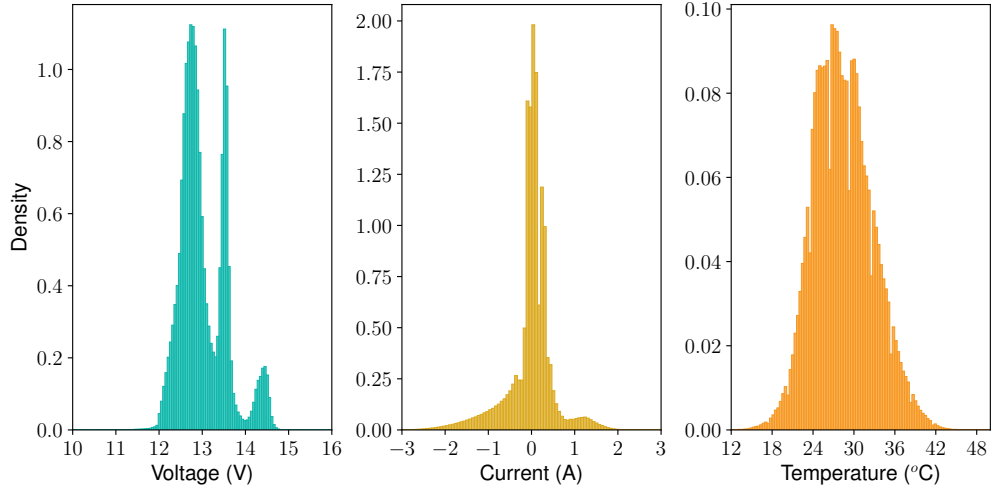

Figure 1: Distributions of voltage, current and temperature in the dataset.

Using the steps outlined above, we obtained a set of 1027 batteries with 491 faulty batteries and 536 healthy ones. This dataset consists of 620 million rows of telemetry data, each row including a timestamp, voltage, current and temperature and the arrival date at the repair shop for each battery where applicable. The total uncompressed size of the dataset (csv format) is 49 GB. The distributions of voltage, current and temperature in the data are shown in Fig. 1. It can be seen that the average currents are small (the nominal capacity is 20 Ah, making the 99th percentile C-rate approximately 0.09C). The temperatures across the population have a reasonably broad range, with the distance between 1st and 99th percentile being approximately 14 °C. The voltage distribution shows 3 peaks, where the two peaks at 13.5 V and 14.5 V are the floating voltage and upper voltage limit during charging respectively. Fig. 2 in turn shows the distribution of battery ages in our dataset, where distribution of ages of healthy batteries reflects that of the overall population, as healthy batteries were sampled using stratification with respect to battery age.

## 2 Determination of open circuit voltage and electrolyte volume

For the Coulomb counting approach for state of charge estimation, described in the Methods section, we required an estimate for the electrolyte volume for the batteries in our dataset. This was obtained by a least squares fit of the OCV curve for lead acid cells from Bode [1] against experimentally measured lab OCV data. First, we carried out a GITT procedure with twenty 1 hour C/20 discharge segments, each with a 2 hour rest in-between them, at 25 °C, to retrieve pointwise estimates of the open circuit voltage as a function of capacity (see Fig. 3).

To determine the electrolyte volume, we note that the change in electrolyte concentration over discharge is given by

$$\Delta c_{HSO_4^-} = \frac{I \Delta t}{V_{elec} F},$$

where $V_{elec}$ is the electrolyte volume and $F$ is Faraday's constant. Bode expresses the OCV of a single cell as a function of acid molality $m$,

$$V_{cell}(m) = 1.922 + 0.148 \log_{10}(m) + 0.064 \log_{10}^2(m) + 0.074 \log_{10}^3(m) + 0.034 \log_{10}^4(m), \quad (1)$$

2

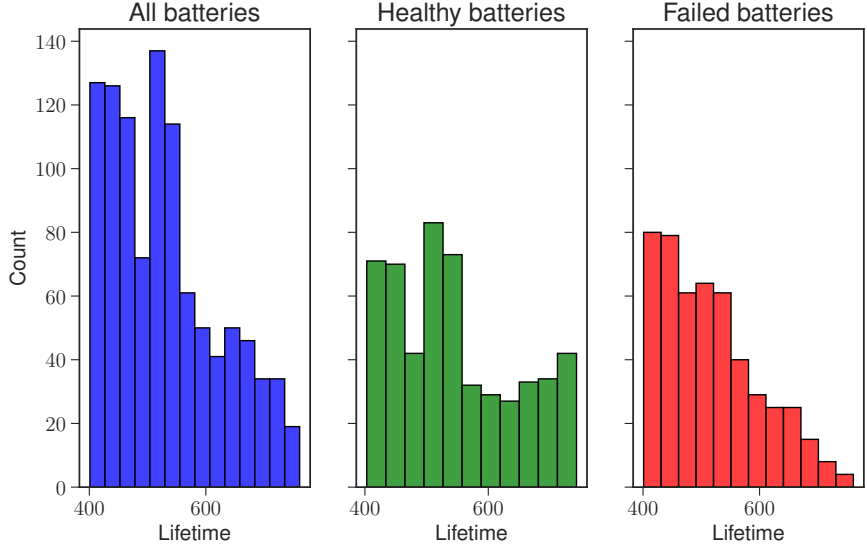

Figure 2: Histograms indicating the age distribution of the overall data as well as the healthy and failed (repaired) sets.

so we made the simple conversion from molality to molarity [2],

$$m = \frac{V_w c_{HSO_4^-}}{(1 - c_{HSO_4^-} V_e) M_w},$$

where $c_{HSO_4^-}$ is acid molarity, $V_w$, $V_e$ and $M_w$ the partial molar volumes of water and acid, and the molar mass of water respectively. (Note that 12 V lead-acid batteries actually contain six cells internally in series, $V_{cell}$ is the voltage of one of these, so $V_0 = 6 V_{cell}$.) We then were able to express our experimentally measured OCV curve (Fig. 3), which was originally a function of Coulombs, as a function of acid concentration instead. To find the best-fit value of $V_{elec}$ we first inverted the Bode curve at a point in the middle on the range illustrated in Fig. 3, then used a least squares fit over the range shown to match the gradient, which is inversely proportional to $V_{elec}$. The optimal value for $V_{elec}$ was found to be 143 cm$^3$.

For further use in modelling, we parameterized the experimental OCV obtained through GITT using a cubic polynomial as a function of estimated acid molarity. As the measured temperature range we observed in the dataset was approximately 10 K and the sensitivity of the OCV to temperature is on the order of 0.1 mV K$^{-1}$/cell [3], we ignored its temperature dependency.

## 3 Recursive estimation of the function $R_0(t, I_t, T_t, \hat{c}_t)$

To achieve linear computational scaling over $n$ data rows when estimating the posterior distribution of $R_0$ for each battery, we employed the recursive framework pioneered by Särkkä et al. [4], [5]. This relies on the interpretation of a Gaussian Process as the solution of a linear stochastic differential equation, where the spectral density of the noise process is described by the covariance function of the GP in the frequency domain. As a result, the GP may be described by a linear dynamic system of the type

$$\frac{\partial R_0(t, I, T, \hat{c})}{\partial t} = F R_0(t, I, T, \hat{c}) + L w(t, I, T, \hat{c}), \qquad (2)$$

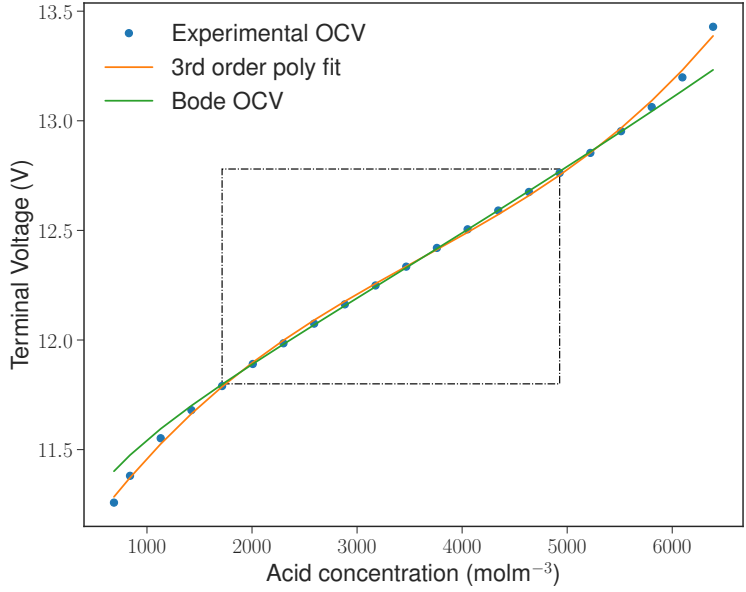

Figure 3: Experimentally determined point-wise estimates of OCV determined by GITT at 25 °C and a 3rd order polynomial fit vs. lead-acid OCV of Bode [1]. We matched the gradients by least squares over a range of acid concentrations in the range shown (central box).

where the transition matrix $F$, dispersion matrix $L$ and the spectral density of the noise term $w(t, I, T, \hat{c})$ are parameterised by the hyperparameters of the GP kernel function. Here, the posterior of $R_0$ and the so-called energy function (the negative logarithm of the unnormalized posterior probability of the hyperparameters) may be obtained by employing standard Kalman filtering and smoothing techniques [6]. Fig. 4 illustrates the concept.

Recall that the kernel function of the GP is given by

$$k(I, T, \hat{c}, t) = \sigma_{f,0}^2 \left( \frac{\min^3(t, t')}{3} + |t - t'| \frac{\min^2(t, t')}{2} \right) + \sigma_{f,1}^2 \exp \left( \sum_{x \in \{I,T,\hat{c}\}} -\frac{(x - x')^2}{2l_x^2} \right). \tag{3}$$

As shown by Solin [7], the solution to the stochastic equation 2 (i.e. the mean and covariance of the solution trajectories) is equivalent to the posterior distribution of the Gaussian process. We find this using a (discrete time) Kalman filter, propagating the mean and covariance of the linear dynamic system using

$$z_{k+1} = A_{\text{GP}} z_k \tag{4a}$$
$$P_{k+1} = A_{\text{GP}} P_k A_{\text{GP}}^\mathsf{T} + Q_{\text{GP}}, \tag{4b}$$

with state vector $z$ being a concatenation of the two processes,

$$z = [z_{\text{GP},0}\ z_{\text{GP},1}]^T, \tag{5}$$

where $z_{\text{GP},0}$ represents the degradation GP and consists of the vector $[R_t, \partial R_t/\partial t]^T$ and $z_{\text{GP},1}$ represents the GP over operating points, which is constant over time. Equation (2) is infinite-dimensional, so we use a finite dimensional representation over the operating points by choosing, via k-means clustering, 20 inducing points over $(I, T, \hat{c})$ for each battery, making $z_{\text{GP},1} \in \mathbb{R}^{20}$. Given

4

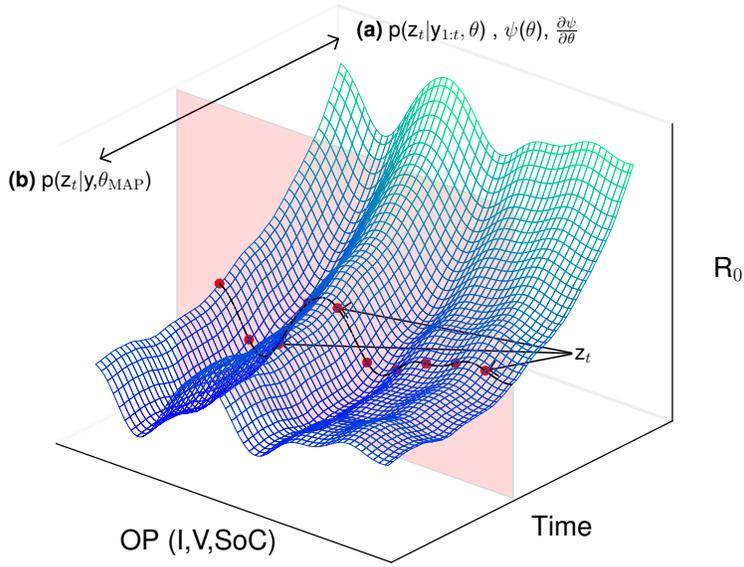

Figure 4: Fitting a GP recursively allows for linear scaling in computational effort in the time axis. State vector $z_t$ represents the GP value at time $t$ at 20 specific operating points, unique to each battery, which were chosen using the k-means algorithm. These 20 points were propagated and updated using a Kalman filter. A gradient based optimizer finds MAP estimates of the hyperparameters by repeatedly running the model 'forwards', (a), to evaluate the energy function and its gradient with respect to hyperparameters. The final posterior distribution is smoothed with a backward pass, (b).

the state vector, the discrete time transition matrix $A_{\text{GP}}$ is block diagonal,

$$A_{\text{GP}} = \text{expm}\left(\begin{bmatrix} F_{\text{GP},0} & 0 \\ 0 & F_{\text{GP},1} \end{bmatrix} \Delta t\right), \text{ where } F_{\text{GP},0} = \begin{bmatrix} 0 & 1 \\ 0 & 0 \end{bmatrix}, F_{\text{GP},1} = \mathbf{0}, \quad (6)$$

where $\mathbf{0}$ indicates a square matrix of zeros of size 20 and $\Delta t$ denotes the (normalised) time-step between charging segments. The process covariance $Q_{\text{GP}}$ is also block-diagonal, given by

$$Q_{\text{GP}} = \begin{bmatrix} Q_{\text{GP},0} & 0 \\ 0 & Q_{\text{GP},1} \end{bmatrix}, \text{ where } Q_{\text{GP},0} = \begin{bmatrix} \frac{1}{3}\Delta t^3 & \frac{1}{2}\Delta t^2 \\ \frac{1}{2}t^2 & \Delta t \end{bmatrix}, Q_{\text{GP},1} = \mathbf{0} \quad (7)$$

As we use a zero-mean GP, the state vector $z_{\text{GP}}$ was initialized at zero. The covariance matrix for the dynamic system is likewise block-diagonal,

$$P_{\text{GP}} = \begin{bmatrix} P_{\text{GP},0} & 0 \\ 0 & P_{\text{GP},1} \end{bmatrix} \quad (8)$$

initialised separately for the two subsystems. The integrated Wiener process is defined to have zero initial covariance [7], so that $P_{\text{GP},0} = \mathbf{0}$. For the second GP over operating points, which has no time dependency, the initial covariance is given by $P_{\text{GP},1} = \sigma_{f,1}^2 \mathbf{I}$, where $\mathbf{I}$ is the identity matrix of size 20 due to the number of inducing points chosen.

5

We can, equivalently, express the propagation of the states and covariances in the form of a joint distribution of previous posterior and current prior states, $z_{t-1}^+$ and $z_t^-$ respectively, given by [5]

$$p\left(\begin{bmatrix} z_{t-1}^+ \\ z_t^- \end{bmatrix}\right) \sim \mathcal{N}\left(\begin{bmatrix} m_{t-1}^+ \\ A_{GP} m_t^- \end{bmatrix}, \begin{bmatrix} P_{t-1}^+ & P_{t-1}^+ A_{GP}^T \\ A_{GP} P_{t-1}^+ & A_{GP} P_{t-1}^+ A_{GP}^T + Q_{GP} \end{bmatrix}\right), \quad (9)$$

where $m_t^-$ and $m_{t-1}^+$ denote the current prior and previous posterior means. The observation model contains the GP describing the dependency on operating point. This means that the estimated internal resistance at any point in time may be expressed as a linear combination of the elements within state vector $z_t$. The resulting observation matrix $H_{GP,t}$ is determined by the Gaussian process over operating points. Summarising, observation model is given by the following distributions [8],

$$z_t \sim \mathcal{N}(m_t^-, P_t^-) \quad (10a)$$
$$R_{t,0}|z_t \sim \mathcal{N}(H_{GP,t} z_t^-, \sigma_{GP,t}^2) \quad (10b)$$
$$V_t|R_{0,t} \sim \mathcal{N}(V_0(\hat{c}_t) + R_{0,t} I_t, \sigma_{n,t}^2). \quad (10c)$$

This gives the joint distribution

$$p\left(\begin{bmatrix} z_{t-1} \\ R_{t,0} \\ V_t \end{bmatrix}\right) \sim \mathcal{N}(m', P'), \text{ where} \quad (11a)$$

$$m' = \begin{bmatrix} m_t^- \\ H_{GP} m_t^- \\ V_0(\hat{c}_t) + H_{GP} m_t^- I_t \end{bmatrix}, \quad (11b)$$

$$P' = \begin{bmatrix} P_t^- & P_t^- H_{GP,t}^T & P_t^- H_{GP,t} I_{d,t}^T \\ H_{GP,t} P_t^- & H_{GP,t} P_t^- H_{GP,t}^T + \sigma_{GP,t}^2 & \left[H_{GP,t} P_t^- H_{GP,t}^T + \sigma_{GP,t}^2\right] I_{d,t}^T \\ I_{d,t} H_{GP,t} P_t^- & I_{d,t} \left[H_{GP,t} P_t^- H_{GP,t}^T + \sigma_{GP,t}^2\right] & \mathbf{I_{d,t} \left[H_{GP,t} P_t^- H_{GP,t}^T + \sigma_{GP,t}^2\right] I_{d,t}^T + \sigma_{n,t}^2} \end{bmatrix}. \quad (11c)$$

The innovation covariance of the Kalman filter is the last block element on the leading diagonal (highlighted in bold). The matrix $I_{d,t}$ is a diagonal matrix consisting of the current at each point of the charging segment. The matrix $H_{GP,t} \in \mathbb{R}^{n \star 22}$, $n$ being the number of data points in the charging segment, is given by

$$H_{GP,t} = \begin{bmatrix} H_0 & H_1 \end{bmatrix} = \begin{bmatrix} 1 & 0 & k(x_t, X_u) \left[k(X_u, X_u) + \text{diag}[P_{GP,1,t}^-]\right]^{-1} \\ \vdots & \vdots & \vdots \end{bmatrix}, \quad (12)$$

where $k(.)$ is the SE kernel covariance function in (3), $x_t = [I_t \; T_t \; \hat{c}_t]$, $X_u \in \mathbb{R}^{20 \star 3}$ is the location of the inducing points in terms of $(I_t, T_t, \hat{c}_t)$ and diag[.] denotes the operator returning the leading diagonal of a matrix. Note, we add the diagonal of the prior state covariance matrix estimate $P_{GP,1,t}^-$ to $k(X_u, X_u)$ in (12) to reflect the uncertainty in the states as they are propagated through time. The uncertainty around the points can vary significantly due to the observed $(I_t, T_t, \hat{c}_t)$ changing from one segment to the next. The variance in the GP is given by

$$\sigma_{GP,t}^2 = \text{diag}\left[k(x_t, x_t) - k(x_t, X_u) \left[k(X_u, X_u) + \text{diag}\left[P_{GP,1,t}^-\right]\right]^{-1} k(X_u, x_t)\right], \quad (13)$$

6

and the calculation of $\sigma_{n,t}^2$, which takes into account the variance in $V_0(\hat{c}_t)$, is described in the Methods section. To calculate maximum-a-posteriori (MAP) estimates of the hyperparameters, we employ the standard approach via Bayes' theorem, noting that the denominator is independent of the hyperparameters,

$$p(\theta_h|D) = \frac{p(D|\theta_h)p(\theta_h)}{p(D)} \propto p(D|\theta_h)p(\theta_h). \tag{14}$$

where $\theta_h = \begin{bmatrix} \sigma_{f,0} & \sigma_{f,1} & l_x \end{bmatrix}$ is a vector of hyperparameters and $D$ represents data, consisting of the input-output pairs through each charging segment (current, state of charge, temperature being the inputs, terminal voltage the output).

For each pass through the data, pointwise estimates of the posterior probability $p(\theta_h|D)$ are given by the energy function, which can be included as part of the recursive Kalman filter approach. The energy function is [5]

$$\Phi(\theta_h) = -\log p(D|\theta) - \log p(\theta) = -\log p(\theta_h) + \frac{1}{2}\sum_t \log|2\pi S_t(\theta_h)| + \frac{1}{2}\sum_t e_t^T S_t(\theta_h)^{-1} e_t, \tag{15}$$

where $p(\theta_h)$ is the hyperprior specified in the Methods section, $e_t$ the error vector for each charging segment, and $S_t(\theta_h)$ the innovation covariance for the segment, given by the lower right diagonal element in equation 11c. This was minimised to find the MAP estimate of $\theta_h$ using the L-BFGS-B algorithm [9], implemented in SciPy. To improve efficiency, the Jacobian of the energy function was determined analytically by extending the recursive method outlined by Mbalawata et al. [10] to include terms for the observation model, equations 10-13. Differentiating the energy function with respect to $\theta_h$ and dropping the dependency on $\theta_h$ in notation for convenience, we get

$$\begin{aligned}
\frac{\partial \Phi}{\partial \theta_h} = &- \frac{\partial \log p(\theta_h)}{\partial \theta_h} + \frac{1}{2}\sum_n \mathrm{Tr}\left(S_t^{-1}\frac{\partial S_t}{\partial \theta_h}\right) \\
&- \frac{1}{2}\sum_n \left(H_t\frac{\partial z_t^-}{\partial \theta_h} + \frac{\partial H_t}{\partial \theta_h}z_t^-\right)^T S_t^{-1} e_t \\
&- \frac{1}{2}\sum_n e_t^T S_t^{-1}\frac{\partial S_t}{\partial \theta_h}S_t^{-1} \\
&- \frac{1}{2}\sum_n e_t^T S_t^{-1}\left(H_t\frac{\partial z_t^-}{\partial \theta_h} + \frac{\partial H_t}{\partial \theta_h}z_t^-\right),
\end{aligned} \tag{16}$$

where Tr denotes the matrix trace and

$$\frac{\partial S_t}{\partial \theta_h} = \frac{\partial H_t}{\partial \theta_h}P_t^- H_t^T + H_t\frac{\partial P_t^-}{\partial \theta_h}H_t^T + H_t P_t^-\left(\frac{\partial H_t}{\partial \theta_h}\right)^T + \frac{\partial \sigma_{r,t}^2}{\partial \theta_h}. \tag{17}$$

Given equations 16–17, the standard Kalman filter states can be augmented with their partial derivatives with respect to the hyperparameters, $\partial z_t/\partial \theta_h$, $\partial P_t^-/\partial \theta_h$, and propagated as described by Mbalawata et al. [10], using matrix fraction decomposition for the propagation of the augmented covariance. The downside of this approach is the repeated calculation over discharge segments of the large exponential of the matrix $M_t \in \mathbb{R}^{88*88}$,

$$\mathrm{expm}(M_t) = \mathrm{expm}\left(\begin{bmatrix} F(\theta_h) & \Sigma(\theta_h) & \mathbf{0} & \mathbf{0} \\ \mathbf{0} & -F^T(\theta_h) & \mathbf{0} & \mathbf{0} \\ G(\theta_h) & W(\theta_h) & F(\theta_h) & \Sigma(\theta_h) \\ \mathbf{0} & -G^T(\theta_h) & \mathbf{0} & -F^T(\theta_h) \end{bmatrix}\Delta t\right), \tag{18}$$

7

where $F(\theta_h)$ is the continuous time transition matrix of the state space system as defined by (6), $G(\theta_h) = \partial F(\theta_h)/\partial \theta_h$, and $W(\theta_h) = \partial \Sigma(\theta_h)/\partial \theta_h$, where $\Sigma$ is the continuous time diffusion matrix of the GP. However, the additive kernel (3) has the convenient property that the matrix exponential (18) has an exact solution for each element in the hyperparameter vector $\theta_h$, foregoing the need to evaluate the matrix exponential using standard numerical methods, therefore requiring a fraction of the computational cost. As a result, the dominant cost in the Kalman filter recursion becomes the inversion of the innovation covariance, $S_t$, which is of the order 100x100 for each discharge segment. This is already incorporated in the standard recursion, making the analytical method for evaluating the energy function gradient approximately 30% faster (and more stable) than the numerical equivalent.

Given the MAP estimates of the hyperparameters, the posterior of $R_0(x)$ was then obtained by using the RTS smoother [11]. This posterior could be evaluated to extrapolate the value of $R_0$ to any given operating point by using

$$R_0(x) \sim \mathcal{N}(\mu_p, \sigma_p^2) \tag{19a}$$

$$\mu = k(x, X_u)\left[k(X_u, X_u) + \text{diag}\left[P_{GP,1}^s\right]\right]^{-1} z_{GP,1}^s, \tag{19b}$$

$$\sigma_p^2 = k(x, x) - k(x, X_u)\left[k(X_u, X_u) + \text{diag}[P_{GP,1}^s]\right]^{-1} k(X_u, x) \tag{19c}$$

where $z_{GP,1}^s$ and $P_{GP,1}^s$ denote the RTS smoothed mean and covariances, which are **constant** as a function of time for the GP over the operating point. For the degradation process, the value of $z_{GP,0}$ and $P_{GP,0}$ can be obtained by applying the forward propagating equations (9) using as a starting point the RTS smoothed values.

The method used for benchmarking our approach is similar to this, but because $R_0$ was considered to be independent of the operating point and modelled by a random walk through time, the state vector $z_t$, variance $P_t$ and process variance $Q$ become scalar, and transition matrix $A_{GP}$ is simply $I$ (in fact 1, since in this case the state is a scalar). The vector valued observations are the same and the measurement error was the fixed value $I\sigma_{n,t}^2$, as in the main approach. The only hyperparameter $Q$ was estimated by maximum likelihood using the same recursion technique as the main method. The initial variance of the state vector was set to $10Q$ in each case.

## 4 Estimated hyperparameters of $R_0$

We calculated the MAP estimates of the hyperparameters for the kernel function, equation 3, for each battery in our dataset using a parallel approach. Hyperparameters were estimated once using the 'zero' time horizon dataset (i.e. the longest dataset). These hyperparameters were then used to estimate the posterior distributions in all cases. Fig. 5 shows the distribution of the hyperparameter MAP estimates and their median estimates in each case. The length scale $l_l$ and magnitudes $\sigma$ have relatively tight distributions, where as the temperature and SoC length scales have some outliers. Also, the median length scales indicate that the applied current is the most significant input, which is consistent with the results showing the largest range in estimated $R_0$ across this input dimension.

## 5 Calculation of stress factors for GP classifier

The augmented input matrix for the GP classifier for fault diagnosis/prognosis consisted of $R_0$, $\partial R_0/\partial t$ and cumulative stress factors known to affect battery state of health. Firstly, battery age

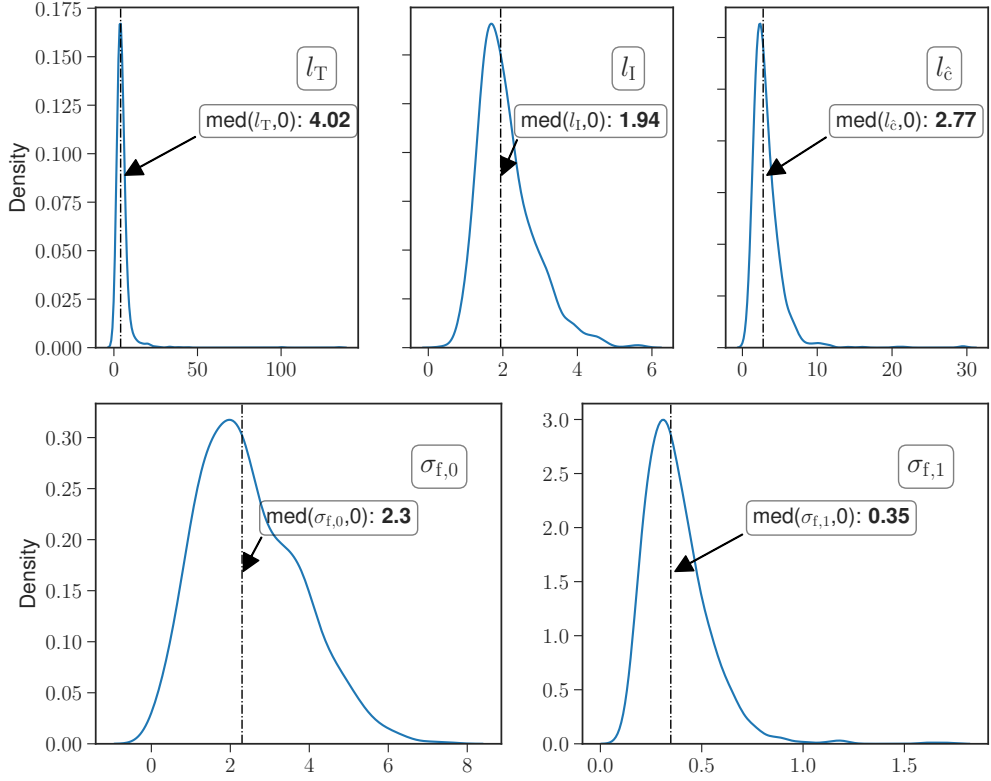

Figure 5: Distributions of each hyperparameter across the population. The median is shown for each hyperparameter.

was defined as the time (in days) since the unit was marked as *activated* by BBOXX. Cumulative time at float charge was calculated by simply summing time increments conditional on the battery being in the float charge region, so that

$$\Sigma_S = \sum_t S(t)\Delta t \quad \text{where } S = \begin{cases} 1 & \text{if in float charge} \\ 0 & \text{otherwise} \end{cases}, \tag{20}$$

where we define the float charge region as a period of over 600 seconds where the voltage is between 13.4 V and 13.6 V. We also calculated the total Ah throughput by taking the absolute cumulative Coulomb count through battery lifetime, and the mean temperatures and voltages by using an expanding window from beginning of life. Finally, we calculated the discharge cycle count by adding up the discharge segments defined by any continuous discharge period of over 600 s where I>0.05 A.

To investigate how many stress factors to include, we calculated the correlation between the health estimates $R_0$ and $\partial R_0/\partial t$ and the stress factors introduced above. Fig. 6 shows the resulting Pearson correlation coefficients calculated at the end of the data series for each battery, for the entire population. Although the correlation coefficient does not capture non-linear interactions between features, it is clear that the health metrics $R_0$ and $\partial R_0/\partial t$ are strongly related, which implies the existence of a degradation knee-point in many of the SoH trajectories. Other factors are generally not strongly correlated, indicating that a wide range of different usage conditions are explored in the data. This is an important advantage of learning battery degradation behaviour from

9

large field datasets—creating a comparably wide variety of operating conditions in a laboratory would be prohibitively expensive and time-consuming. To assess which factors might be more important in causing end of life, we can compare their relative importance for fault prediction, as discussed in section 5 in the main text.

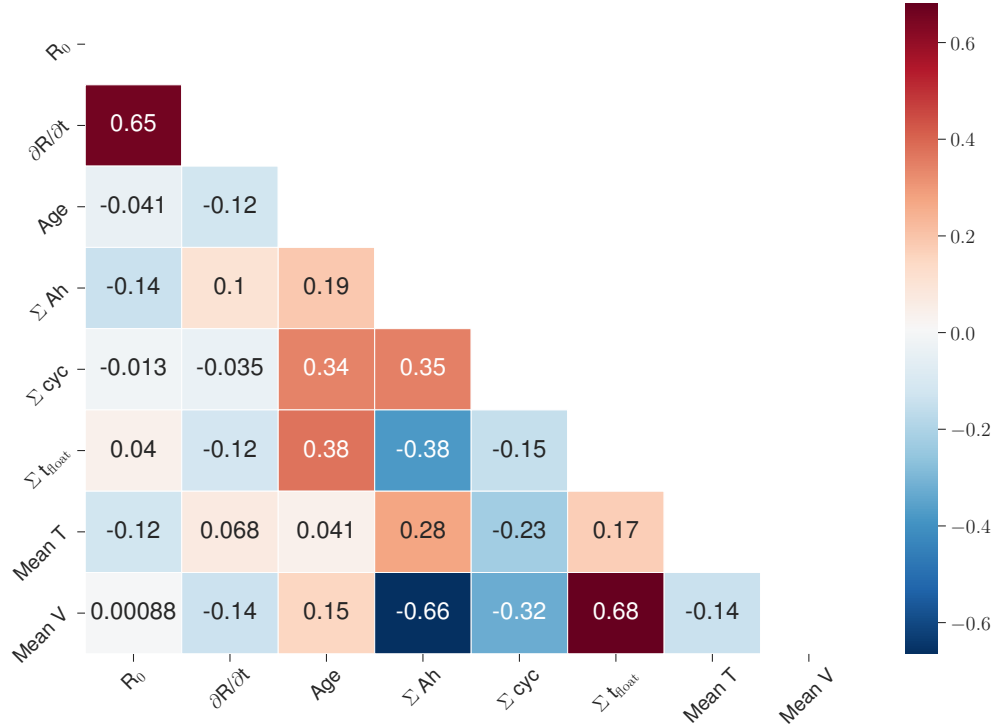

Figure 6: Battery SoH, measured by internal resistance, complements other aging factors, such as calendar age, as shown in this correlation matrix for health metrics $R_0$, $\partial R_0/\partial t$ and stress factors associated with battery degradation. $\Sigma$ denotes the cumulative sum up to the point of diagnosis.

# 6 GP classifier test set and performance matrix

As discussed in Methods, training and testing of the GP classifier was done by nested cross-validation. First, we used 5-fold stratified cross validation to split the 'master' dataset of 1027 batteries with a 48/52% failed/healthy split in to training in test sets. In addition we created extra testing subsets within each test set, which had different proportions of failed vs. healthy batteries. These were created by taking 10 random samples of 40%, 60% and 80% of the existing failed batteries in the test set while keeping the healthy set fixed, making the sub-test sets smaller in size. Altogether we ended up with 3100 tests cases, where

$3100 = (4 \text{ input sets}) * (5 \text{ CV sets}) * (5 \text{ time horizons}) * (10 * 3 \text{ random subsets} + 1 \text{ complete test set})$

The main text summarises the performance by time horizon. The full performance data is shown in Table 1. In each case we show the number of true positives, false positives, false negatives and true negatives (TP, FP, FN, TN respectively) from the classifier predictions in the test set with 52/48% healthy/failed split using 5-fold cross-validation. The balanced accuracy is the average of sensitivity and specificity for each case (see Methods section in main paper). It is clear that

| Days before | Classifier input | TP | FP | FN | TN | Balanced Accuracy (%) |
|---|---|---|---|---|---|---|
| 0 | R, $\frac{\partial R}{\partial t}$ | 339 | 77 | 152 | 459 | 77.3 |
| 0 | R, $\frac{\partial R}{\partial t}$, stress factors | 389 | 77 | 102 | 459 | 82.4 |
| 0 | Benchmark | 244 | 95 | 247 | 441 | 66.0 |
| 0 | Stress factors only | 313 | 120 | 178 | 416 | 70.7 |
| 14 | R, $\frac{\partial R}{\partial t}$ | 308 | 87 | 183 | 449 | 73.2 |
| 14 | R, $\frac{\partial R}{\partial t}$, stress factors | 378 | 88 | 113 | 448 | 80.3 |
| 14 | Benchmark | 203 | 90 | 288 | 446 | 62.3 |
| 14 | Stress factors only | 336 | 139 | 155 | 397 | 71.2 |
| 28 | Stress factors only | 346 | 154 | 145 | 382 | 70.9 |
| 28 | Benchmark | 170 | 126 | 321 | 410 | 55.6 |
| 28 | R, $\frac{\partial R}{\partial t}$, stress factors | 373 | 126 | 118 | 410 | 76.2 |
| 28 | R, $\frac{\partial R}{\partial t}$ | 276 | 113 | 215 | 423 | 67.6 |
| 42 | R, $\frac{\partial R}{\partial t}$ | 258 | 114 | 233 | 422 | 65.6 |
| 42 | R, $\frac{\partial R}{\partial t}$, stress factors | 367 | 135 | 124 | 401 | 74.8 |
| 42 | Benchmark | 124 | 125 | 367 | 411 | 51.0 |
| 42 | Stress factors only | 353 | 181 | 138 | 355 | 69.1 |
| 56 | R, $\frac{\partial R}{\partial t}$ | 245 | 127 | 246 | 409 | 63.1 |
| 56 | R, $\frac{\partial R}{\partial t}$, stress factors | 368 | 155 | 123 | 381 | 73.0 |
| 56 | Benchmark | 107 | 109 | 384 | 427 | 50.7 |
| 56 | Stress factors only | 357 | 214 | 134 | 322 | 66.4 |

Table 1: Performance of GP classifier over all input sets and time horizons for the case where failed batteries constitute 48% of the population.

the combined model including the inputs $R, \partial R \partial t$ as well as the stress factors has the strongest performance in predicting faults.

# 7 Comparison to benchmark

Fig. 7 shows the comparison between $R_0$ timelines calculated using the method proposed in this paper vs. using the benchmark method (e.g. [12]), as discussed above and in the main text. Normalising for operating conditions and using a smooth GP kernel results in much smoother health trajectories over the lifetime of each battery, giving improved fault classification results.

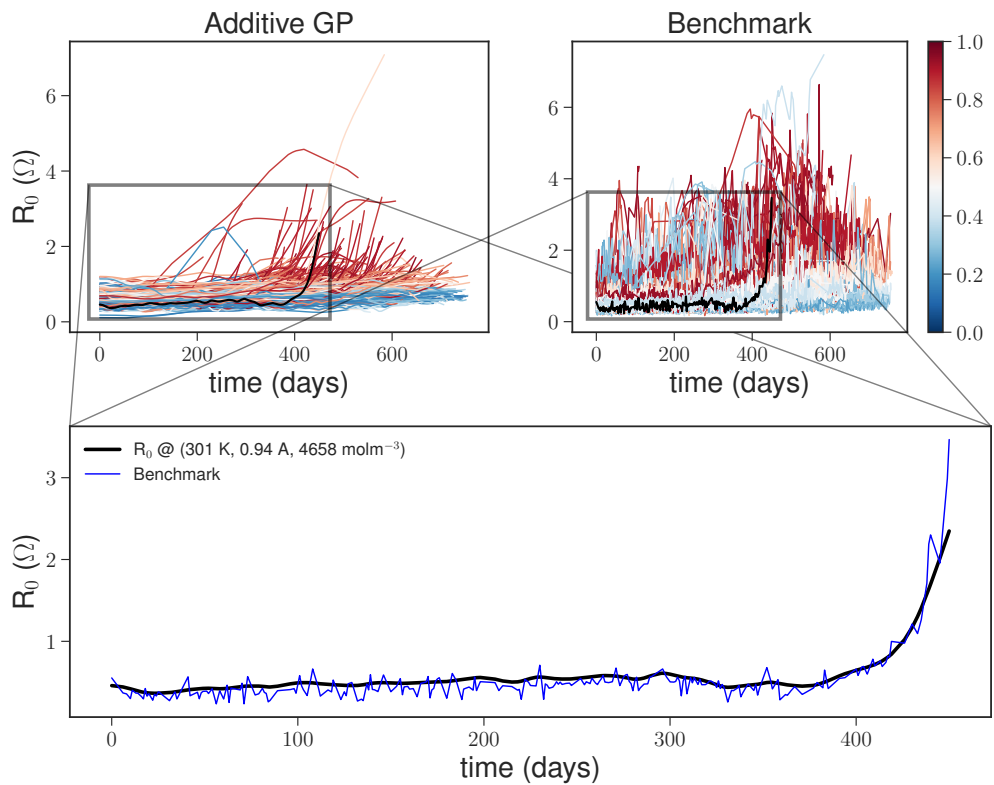

Figure 7: Estimated $R_0$ trajectories using our methods compared with the benchmark approach, illustrated for the population and for a single battery. Our GP method results in much smoother estimates over the lifetime of each battery, by accounting for variations due to operating conditions and imposing smoothness by choice of kernel and prior distribution of hyperparameters.



\end{document}